\documentclass[11pt,a4paper]{article}
\usepackage[hyperref]{emnlp2020}
\usepackage{times}
\usepackage{latexsym}

\usepackage{microtype}

\aclfinalcopy 
\def\aclpaperid{2783} 

\usepackage{algorithm}
\usepackage{algorithmic}
\usepackage{amsmath}
\usepackage{amsfonts}
\usepackage{graphics}
\usepackage{epsfig}
\usepackage{float}
\usepackage{multirow}
\usepackage{appendix}
\usepackage{color}
\usepackage{times}
\usepackage{helvet}
\usepackage{courier}
\usepackage{multirow}
\usepackage{bm}
\usepackage[american]{babel}
\usepackage{graphicx}
\usepackage{enumitem,kantlipsum}
\usepackage{subfig}
\usepackage{xr}

\def \Diag{\text{Diag}}
\def\X{\bm{X}}

\def \v{{\bf v}}
\def \w{{\bf w}}
\def \x{{\bf x}}

\def \z{{\bf z}}
\def \b{{\bf b}}

\def \hw{\hat{\bf{w}}}

\def \W{{\bf W}}

\def \H{{\bf H}}
\def \X{{\bf X}}
\def \I{{\bf I}}
\def \A{{\bf A}}
\def \P{{\bf P}}

\def \K{{\bf K}}
\def \Q{{\bf Q}}

\def \m{{\bf m}}
\def \v{{\bf v}}

\def \v{{\bf v}}

\def \0{{\bf 0}}

\title{TernaryBERT: Distillation-aware Ultra-low Bit BERT}

\author{
	Wei Zhang\thanks{\ \ Authors contribute equally.}, Lu Hou\footnotemark[1], Yichun Yin\footnotemark[1], Lifeng Shang, Xiao Chen, Xin Jiang, Qun Liu
	\\Huawei Noah's Ark Lab
	\\{\{zhangwei379, houlu3, yinyichun, shang.lifeng, chen.xiao2, jiang.xin, qun.liu\}@huawei.com}
}

\date{}

\begin{document}
\maketitle
\begin{abstract}
	Transformer-based pre-training models like BERT have achieved remarkable performance in many natural language processing tasks.
	However, these models are both computation and memory expensive, hindering their deployment to resource-constrained devices.
	 In this work, we propose TernaryBERT,  which ternarizes the weights in a fine-tuned BERT model. Specifically, we use both approximation-based and loss-aware ternarization methods and empirically investigate the ternarization granularity of different parts of BERT. Moreover, to reduce the accuracy degradation caused by the lower capacity of low bits, we leverage the  knowledge distillation  technique~\cite{jiao2019tinybert} in the training process.
Experiments on the GLUE benchmark and SQuAD show that our proposed TernaryBERT outperforms the other BERT quantization methods, and even achieves comparable performance as the full-precision model while being 14.9x smaller.
\end{abstract}

\section{Introduction}
Transformer-based models have shown great power in various natural language processing (NLP) tasks. Pre-trained with gigabytes of unsupervised data, these models usually have hundreds of millions of parameters. 
For instance, the BERT-base model has 109M parameters,
with the model size of 400+MB if represented in 32-bit floating-point format, which is both computation and memory expensive during inference.
This poses great challenges for these models to run on resource-constrained devices like cellphones. 
To alleviate this problem, various methods are proposed to compress these models, like  using low-rank approximation~\citep{ma2019tensorized,lan2020ALBERT}, weight-sharing~\citep{dehghani2018universal,lan2020ALBERT}, knowledge distillation~\citep{sanh2019distilbert,sun2019patient,jiao2019tinybert},
pruning~\citep{michel2019sixteen,voita2019analyzing,fan2019reducing},
adaptive depth and/or width~\citep{liu2020fastbert,hou2020dynabert},
and quantization~\citep{zafrir2019q8bert,shen2019q,fan2020training}.

	\begin{figure}[t!]	
	\centering
	\subfloat{
		\includegraphics[width=0.45\textwidth]{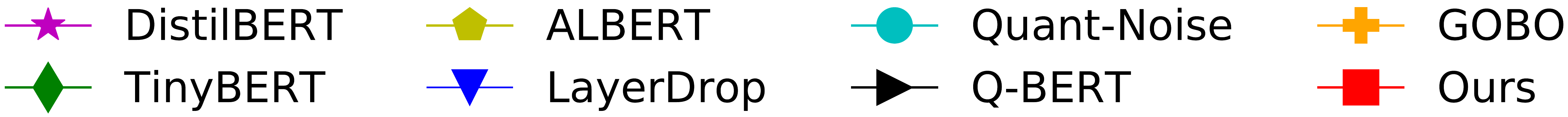}}
	\\
	\addtocounter{subfigure}{-1}
	\includegraphics[width=0.32\textwidth]{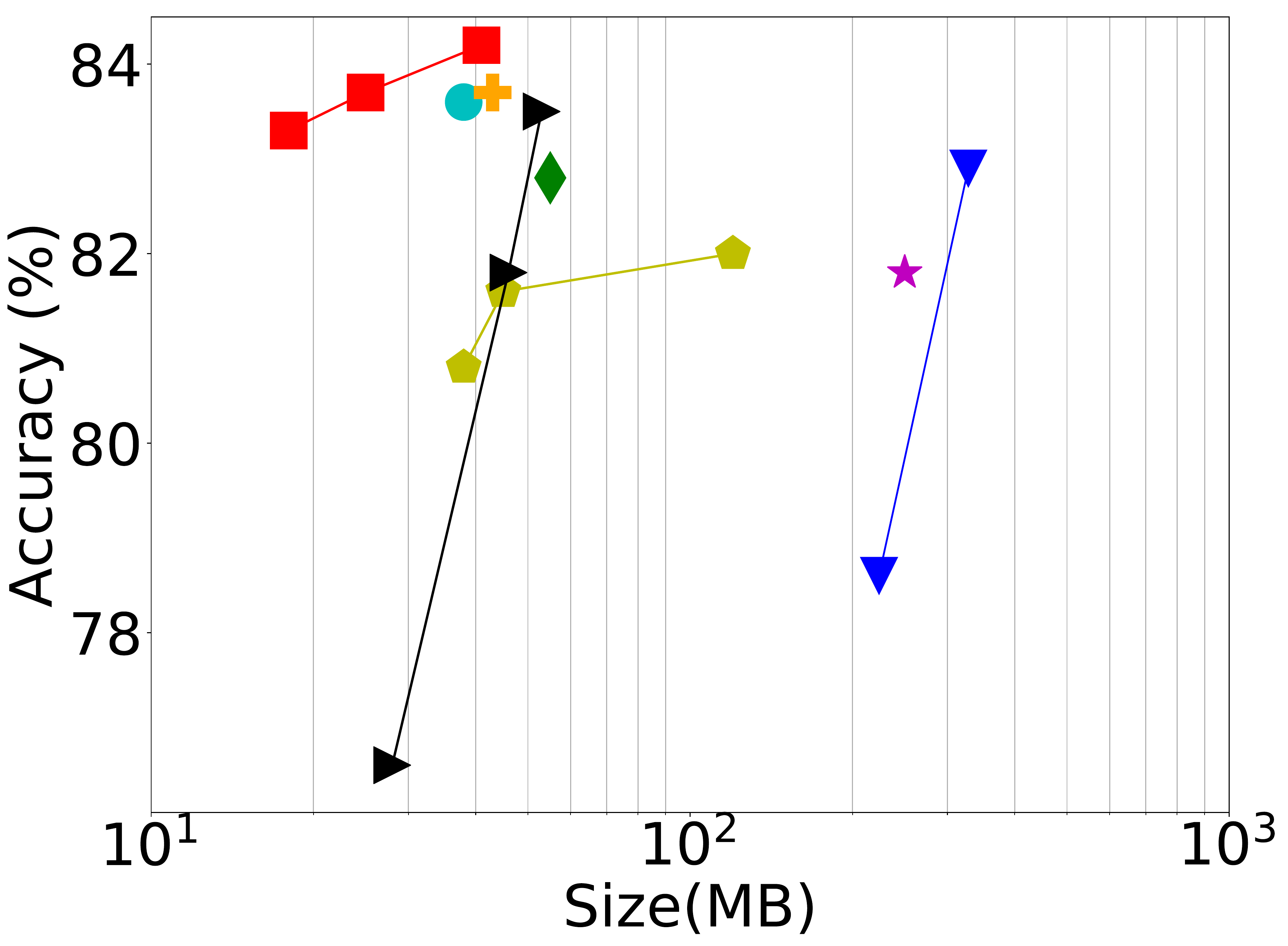}	
	\caption{ 
		Model Size vs. MNLI-m Accuracy. Our proposed method (red squares) outperforms other 
		BERT compression methods. Details are in Section~\ref{sec:comp_other}. }
	\label{fig:comp}
\end{figure}

Compared with other compression methods, quantization compresses a neural network by using lower bits for  weight values without changing the model architecture, and 
is particularly useful for  carefully-designed network architectures like Transformers.
In addition to weight quantization, further quantizing activations can speed up inference with target hardware  by turning  floating-point operations into integer  or bit operations.
In~\citep{prato2019fully, zafrir2019q8bert}, 8-bit quantization is successfully applied to Transformer-based models
with  comparable performance as the full-precision baseline.
However, quantizing these models to ultra low bits (e.g., 1 or 2 bits)  can be much more challenging due to significant reduction in  model capacity.
To avoid severe accuracy drop, more complex quantization methods, like mixed-precision quantization~\citep{shen2019q,zadeh2020gobo} and product quantization (PQ)~\cite{fan2020training}, are used.
However, mixed-precision quantization is unfriendly to some hardwares, 
and PQ requires extra clustering operations.

Besides quantization, 
knowledge distillation 
~\citep{hinton2015distilling}  which transfers knowledge learned in the prediction layer of a cumbersome teacher model to a smaller student model, is also 
widely used to compress BERT~\citep{sanh2019distilbert,sun2019patient,jiao2019tinybert,wang2020minilm}.
Instead of directly being used to compress BERT, the distillation loss can also be used in combination with other compression methods  \citep{mccarley2019pruning,mao2020ladabert,hou2020dynabert},
to fully leverage the knowledge of teacher  model.

In this work, we propose TernaryBERT, whose weights are restricted to $\{-1,0,+1\}$.
Instead of directly using knowledge distillation to compress a model,
 we use it to improve the performance of ternarized student model
 with the same size as the teacher model.
In this way, we wish to transfer the knowledge from the highly-accurate teacher model to the ternarized student model
 with smaller capacity, and to fully explore the compactness by combining quantization and distillation. We investigate the ternarization granularity of different parts of the BERT model, and apply various distillation losses to improve the performance of TernaryBERT.
Figure~\ref{fig:comp} summarizes the accuracy versus model size on MNLI, where our proposed method outperforms
other BERT compression methods. More empirical results on the GLUE benchmark and SQuAD show that our proposed TernaryBERT outperforms other quantization methods, and even achieves 
comparable performance as the full-precision baseline, while being much smaller.

\section{Related Work}
\subsection{Knowledge Distillation}
Knowledge distillation is first proposed in~\citep{hinton2015distilling}  to transfer knowledge in the logits from a large teacher model to a more compact student model  without sacrificing too much performance.
It has achieved remarkable performance 
in NLP~\citep{kim2016sequence,jiao2019tinybert} recently.
Besides the logits~\cite{hinton2015distilling}, knowledge from the intermediate representations~\citep{romero2014fitnets,jiao2019tinybert} and attentions~\citep{jiao2019tinybert,wang2020minilm} are also used to 
guide the training of a smaller BERT.

Instead of directly being used for compression,  
knowledge distillation  can also be used in combination with other compression methods like pruning~\citep{mccarley2019pruning,mao2020ladabert}, low-rank approximation~\citep{mao2020ladabert}
and dynamic networks~\citep{hou2020dynabert},
to fully leverage the knowledge of the teacher BERT model. 
Although combining quantization and  distillation has been explored in convolutional neural networks (CNNs)~\citep{polino2018model,stock2020and,kim2019qkd},
using knowledge distillation to train quantized BERT  has not been studied.
Compared with CNNs which simply perform convolution in each layer, the BERT model is more complicated with each Transformer layer containing both a Multi-Head Attention mechanism and a position-wise Feed-forward
Network. Thus the knowledge that can be distilled in a BERT model is also much richer~\citep{jiao2019tinybert,wang2020minilm}.

\subsection{Quantization}
Quantization has been extensively studied for CNNs.
Popular 
ultra-low bit
weight quantization methods for CNNs  can be  divided into two categories: approximation-based and loss-aware based.
Approximation-based quantization \citep{rastegari2016xnor,li2016ternary} aims at keeping the quantized weights close to the
full-precision weights, while
loss-aware
based quantization~\citep{hou2017loss,hou2018loss,leng2018extremely} directly optimizes for the quantized weights that minimize the training loss.

On Transformer-based models, 8-bit fixed-point quantization is successfully applied in
 fully-quantized Transformer~\citep{prato2019fully} and Q8BERT~\citep{zafrir2019q8bert}.
The use of lower bits is also investigated in \citep{shen2019q,fan2020training,zadeh2020gobo}.
Specifically, 
In Q-BERT~\cite{shen2019q} and GOBO~\citep{zadeh2020gobo}, mixed-precision with 3 or more bits are used to avoid severe accuracy drop.
However, mixed-precision quantization can be unfriendly to some hardwares.
\citet{fan2020training} propose Quant-Noise which quantizes a subset of weights in each iteration
 to allow unbiased gradients to flow through the network. 
Despite the high compression rate achieved, 
the quantization noise rate needs to be tuned for good performance. 

In this work,  we  extend 
	both  approximation-based and loss-aware ternarization methods to different granularities for different parts of the BERT model, i.e., word embedding and weights in Transformer layers.
	To avoid accuracy drop due to the reduced capacity caused by ternarization, various distillation losses are used to guide the training of the ternary  model.  
	
\begin{figure*}[htbp]
	\centering
	\includegraphics[width=0.8\linewidth]{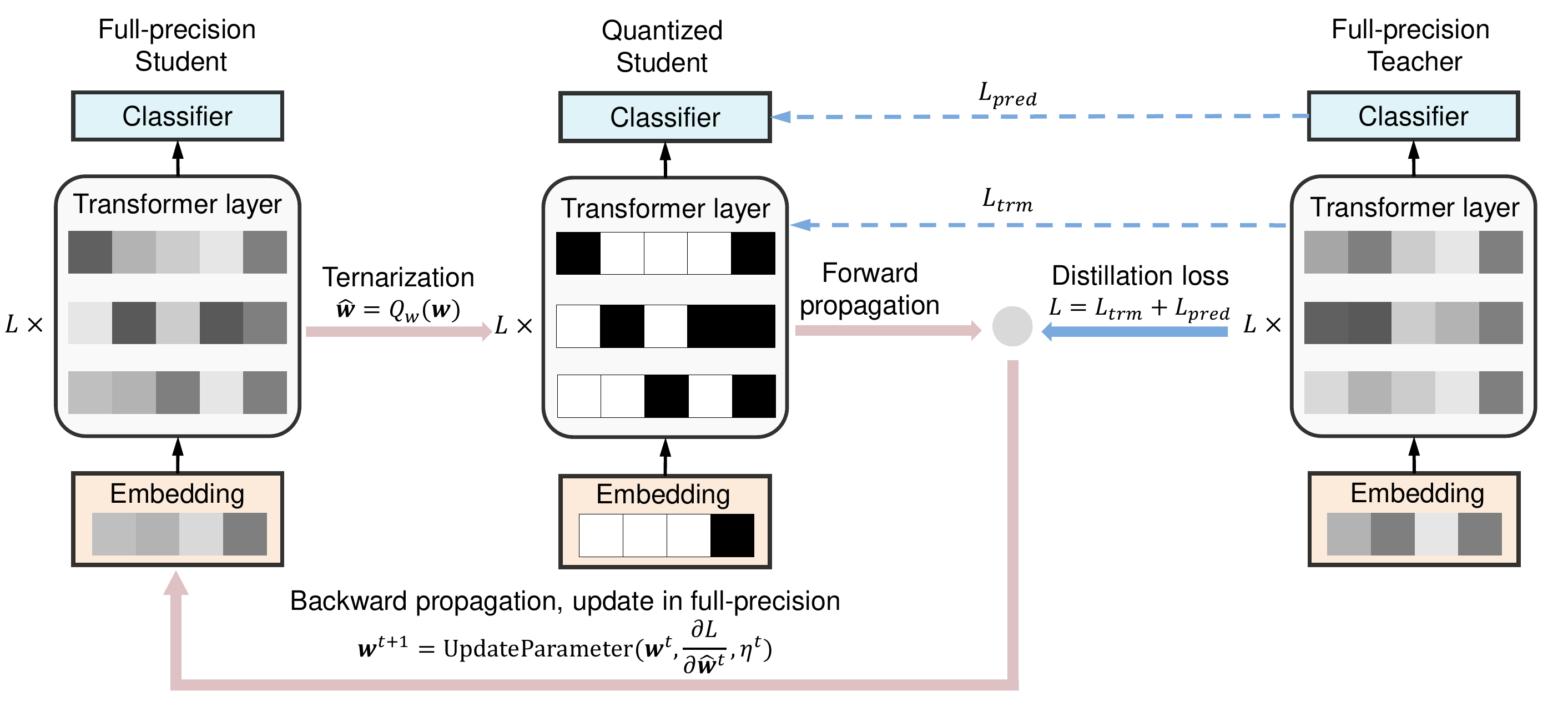}
	\caption{Depiction of the proposed distillation-aware ternarization of BERT model.}
	\label{fig:main}
\end{figure*}

\section{Approach}

In this section, we elaborate on the method of using knowledge distillation to train TernaryBERT, the weights of which take values in $\{-1,0,+1\}$.

Let the full-precision weight in the BERT model be 
$\w$, where
 $\w=\text{vec}(\W)$ 
 returns a vector by stacking all the columns of 
 weight matrix $\W$. The corresponding ternarized weight is
denoted as $\hw = Q_w(\w)$ where $Q_w$ is the weight ternarization function.
The whole framework, which we call Distillation-aware ternarization, is shown in Figure~\ref{fig:main}.
Specifically, at the $t$-th  training iteration, we first ternarize the weights $\w^t$ in the student BERT model to $\hw^t$.
Then we do the forward pass with the ternarized model.
After that, the gradient of the distillation loss w.r.t. the quantized weights $\frac{\partial \mathcal{L}}{\partial \hw^t}$ is computed. As is shown in~\cite{courbariaux2016binarized,hou2018loss}, it is important to keep the full-precision weight during training. 
Hence, we use the full-precision weight for parameter update: $\w^{t+1} = \text{UpdateParameter}(\w^t, \frac{\partial \mathcal{L}}{\partial \hw^t}, \eta^t)$,
where $\eta^t$
is the  learning rate at the $t$-th iteration. 

In the following, we will first  introduce what and how to quantize  in Section~\ref{sec:ternary}.
Then in Section~\ref{sec:distill}, we introduce the distillation loss used to improve the performance of the ternarized model.

\subsection{Quantization}
\label{sec:ternary}

The BERT model~\cite{devlin2019bert} is built with Transformer layers~\cite{vaswani2017attention}. 
A standard Transformer layer includes two main sub-layers: Multi-Head Attention (MHA) module and Feed-Forward Network (FFN).

For the $l$-th Transformer layer, suppose the input to it is 
$\H_l \in \mathbb{R}^{n \times d}$ where $n$ and $d$ 
are the sequence length and hidden state size, respectively.
Suppose there are $N_H$ attention heads in each layer, and head $h$ 
is parameterized by 
$\W^Q_h, \W^K_h, \W^V_h \in\ \mathbb{R}^{d \times d_h}$ 
where $d_h = \frac{d}{N_H}$.
After computing the   attention scores by dot product of queries and keys
\begin{equation}
\A_h = \Q \K^\top = \H_l \W^Q_h \W^{K\top}_h \H_l^\top,
\end{equation}
the softmax function is applied on the normalized scores  to get the  output as
$
\text{head}_h = \text{Softmax}(\frac{1}{\sqrt{d}}\A_h) \H_l \W^V_h.
$
Denote $\W^* = 
[\W^*_1,  \cdots, \W^*_{N_H}]$ 
where $*$ can be $Q,K,V$.
The output of the multi-head attention is:
\begin{eqnarray}
&&\text{MHA}_{\W^Q, \W^K, \W^V,\W^O}(\H_l)  \nonumber\\
&= &\text{Concat}(\text{head}_1, \cdots, \text{head}_{N_H})\W^O. \label{eq:mha}
\end{eqnarray}
The FFN layer composes two linear layers parameterized by   $\W^1 \in  \mathbb{R}^{d \times d_{ff}}, \b^1 \in \mathbb{R}^{d_{ff}}$ and $\W^2 \in \mathbb{R}^{d_{ff}\times d},\b^2\in \mathbb{R}^{d}$ respectively, where $d_{ff}$ is the number of neurons in the intermediate layer of FFN.
Denote  the input to FFN as $\X_l \in \mathbb{R}^{n \times d}$, 
the output is then computed as:
\begin{equation}
\text{FFN}(\X_l) = \text{GeLU}(\X_l\W^1+\b^1)\W^2 + \b^2. \label{eq:ffn}
\end{equation}
Combining (\ref{eq:mha}) and (\ref{eq:ffn}),
the forward propagation for the $l$-th Transformer layer can be written as
\begin{eqnarray*}
	\X_l&=&\text{LN}(\H_l+\text{MHA}(\H_l))\\
	\H_{l+1}&=&\text{LN}(\X_l+\text{FFN}(\X_l)),
\end{eqnarray*}
where $\text{LN}$ is the layer normalization.
The input to the first transformer layer
\begin{equation}
\H_1 =  \text{EMB}_{\W^E,\W^S, \W^P}(\z) \label{eq:embedding}
\end{equation}
is the combination of the token embedding,  segment embedding and  position embedding.
Here $\z$ is the input sequence, and $ \W^E, \W^S, \W^P$ are the 
learnable word embedding, segment embedding and position embedding, respectively.

For weight quantization,
following~\cite{shen2019q,zafrir2019q8bert},
we quantize the weights $\W^Q, \W^K, \W^V, \W^O, \W^1, \W^2$  in  (\ref{eq:mha}) and (\ref{eq:ffn}) from all Transformer layers, as well as the word embedding $\W^E$ in (\ref{eq:embedding}).  
Besides these weights, we also quantize the inputs of all linear layers and matrix multiplication operations in the forward propagation. 
We do not quantize $\W^S, \W^P$, and  the bias in linear layers because the parameters involved are negligible.
Following~\cite{zafrir2019q8bert}, we also do not quantize the softmax operation, layer normalization and the last task-specific layer because the parameters contained in these operations are negligible and 
quantizing them can bring significant accuracy degradation. 
\paragraph{Weight Ternarization.}
In the following, we discuss the choice of the weight ternarization function $Q_w$ in Figure~\ref{fig:main}.
	
Weight ternarization is pioneered in ternary-connect~\cite{lin2016neural} where the ternarized values can take $\{-1,0,1\}$ represented by 2 bits.  By ternarization,
most of the floating-point multiplications in the forward pass are turned into floating-point additions,
which greatly reduces  computation and memory.
Later, by adding a 
 scaling parameter, better results are obtained  in~\cite{li2016ternary}.
Thus in this work, to ternarize the weights of BERT,
we use  
both approximation-based ternarization method TWN~\cite{li2016ternary} and loss-aware ternarization LAT~\cite{hou2018loss}, where the ternary weight
$\hw$ can be represented by the multiplication of a scaling parameter $\alpha>0$ and a ternary vector $\b \in \{-1, 0, +1\}^{n}$ as $\hw= \alpha \b$. Here $n$ is the number of elements in $\hw$.

In the $t$-th training iteration,
TWN 
ternarizes the weights by minimizing the distance between the full-precision weight $\w^t$ and  ternarized weight $\hw^t=\alpha^t\b^t$ with following optimization problem~\citep{li2016ternary}
\begin{eqnarray}
\min_{\alpha^t, \b^t} && \|\w^t - \alpha^t \b^t\|_2^2  \nonumber\\
\text{s.t.} &&   \alpha^t >0, \b^t \in \{-1,0, 1\}^{n}. 	\label{eq:twn}
\end{eqnarray}
Let $\I_{\Delta}(\x)$ be a thresholding function that
$[\I_{\Delta}(\x)]_i\!\!=\!\!1$  if $x_i\!>\!\!\Delta$,  $-1$ if $x_i\!\!<\!\!-\Delta$, and 0 otherwise, where $\Delta$ is a positive threshold.
Let $\odot$ be element-wise multiplication,
the optimal solution of (\ref{eq:twn}) satisfies~\cite{hou2018loss}:
$
 \b^t=\I_{\Delta^t}(\w^t)$ and  $\alpha^t =\frac{\|\b^t\odot \w^t\|_1}{\|\b^t\|_1} ,
 $
where
\begin{equation*}
	\Delta^t \!=\!\arg \max_{\Delta>0} \frac{1}{\|\I_{\Delta}(\w^t)\|_1} \left(\sum_{i :
		|[\w^t]_i|>
		\Delta}
	|[\w^t]_i| \right)^2.
\label{eq:twn1}
\end{equation*}
The exact solution of $\Delta^t$
  requires an expensive sorting operation~\citep{hou2017loss}.
Thus in \citep{li2016ternary}, TWN approximates  the  threshold with $\Delta^t= \frac{0.7\|\w^t\|_1}{n}$. 

Unlike TWN, LAT 
directly searches for the ternary weights that minimize the training
 loss  $\mathcal{L}$.
The ternary weights are obtained by solving the optimization problem:
\begin{eqnarray}
\min_{\alpha,\b} && \mathcal{L}(\alpha \b) \nonumber \\
\text{s.t.} && \alpha > 0, \b \in \{-1,0, 1\}^{n}. 	\label{eq:lat}
\end{eqnarray}
For a vector $\x$, let 
$\sqrt{\x}$ be the element-wise square root,
$\text{Diag}(\x)$ returns a diagonal matrix with $\x$ on the diagonal, and $\|\x\|_Q^2\!\!=\!\!\x^\top Q\x$.
Problem (\ref{eq:lat}) can be reformulated as solving the following sub-problem at the $t$-th iteration~\citep{hou2018loss}
\begin{eqnarray}
\min_{\alpha^t, \b^t} && \|\w^t - \alpha^t \b^t\|_{\Diag(\sqrt{\v^t})}^2  \nonumber\\
\text{s.t.} &&  \alpha^t >0, \b^t \in \{-1,0, 1\}^{n}, 	\label{eq:lat_solution}
\end{eqnarray}
where
$\v^t$ is a diagonal approximation of the Hessian of $\mathcal{L}$
readily available as the second moment of gradient in adaptive learning rate optimizers like 
 Adam~\citep{kingma2014adam}.
Empirically, we use the second moment in BertAdam\footnote{\url{https://github.com/huggingface/transformers/blob/v0.6.2/pytorch_pretrained_bert/optimization.py}},
which is a variant of Adam by fixing the weight decay~\citep{loshchilov2019decoupled} and removing the bias compensation~\citep{kingma2014adam}.
	For (\ref{eq:lat_solution}),
	both an expensive exact solution based on  sorting operation, and an efficient approximate solution based on alternative optimization are provided in \citep{hou2018loss}. In this paper, we use the more efficient approximate solution.

In the original paper of TWN and LAT, one scaling parameter is used for each convolutional or fully-connected layer. In this work, we extend them to the following two  granularities:
(i)	\textbf{layer-wise ternarization} which uses one scaling parameter for  all elements in each weight matrix; and
(ii) \textbf{row-wise ternarization} which uses one scaling parameter for each row in a weight matrix.
With more scaling parameters, row-wise ternarization has finer granularity and smaller quantization error.

\paragraph{Activation Quantization.}
To  make the most expensive matrix multiplication operation faster,
following~\cite{shen2019q,zafrir2019q8bert},
 we also quantize the  activations (i.e., inputs of all linear layers and matrix multiplication) to 8 bits.
There are two kinds of commonly used 8-bit quantization methods: symmetric 
 and min-max 8-bit quantization.
The quantized values  of the symmetric 8-bit quantization  distribute symmetrically in both sides of 0, while 
those of min-max 8-bit quantization distribute uniformly in a range determined by the minimum and maximum values.

\begin{figure}
	\centering
	\includegraphics[width=0.5\linewidth]{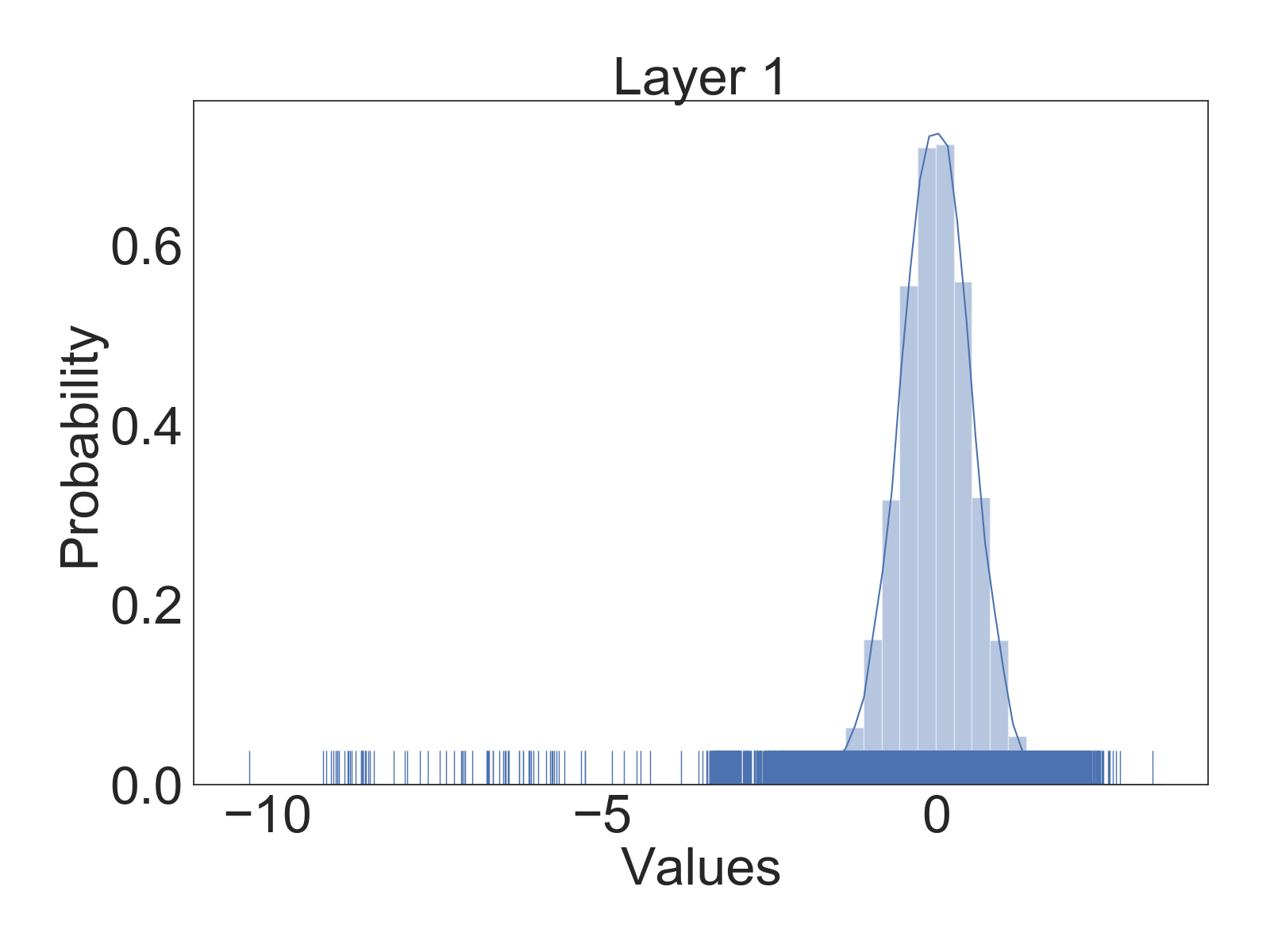}\includegraphics[width=0.5\linewidth]{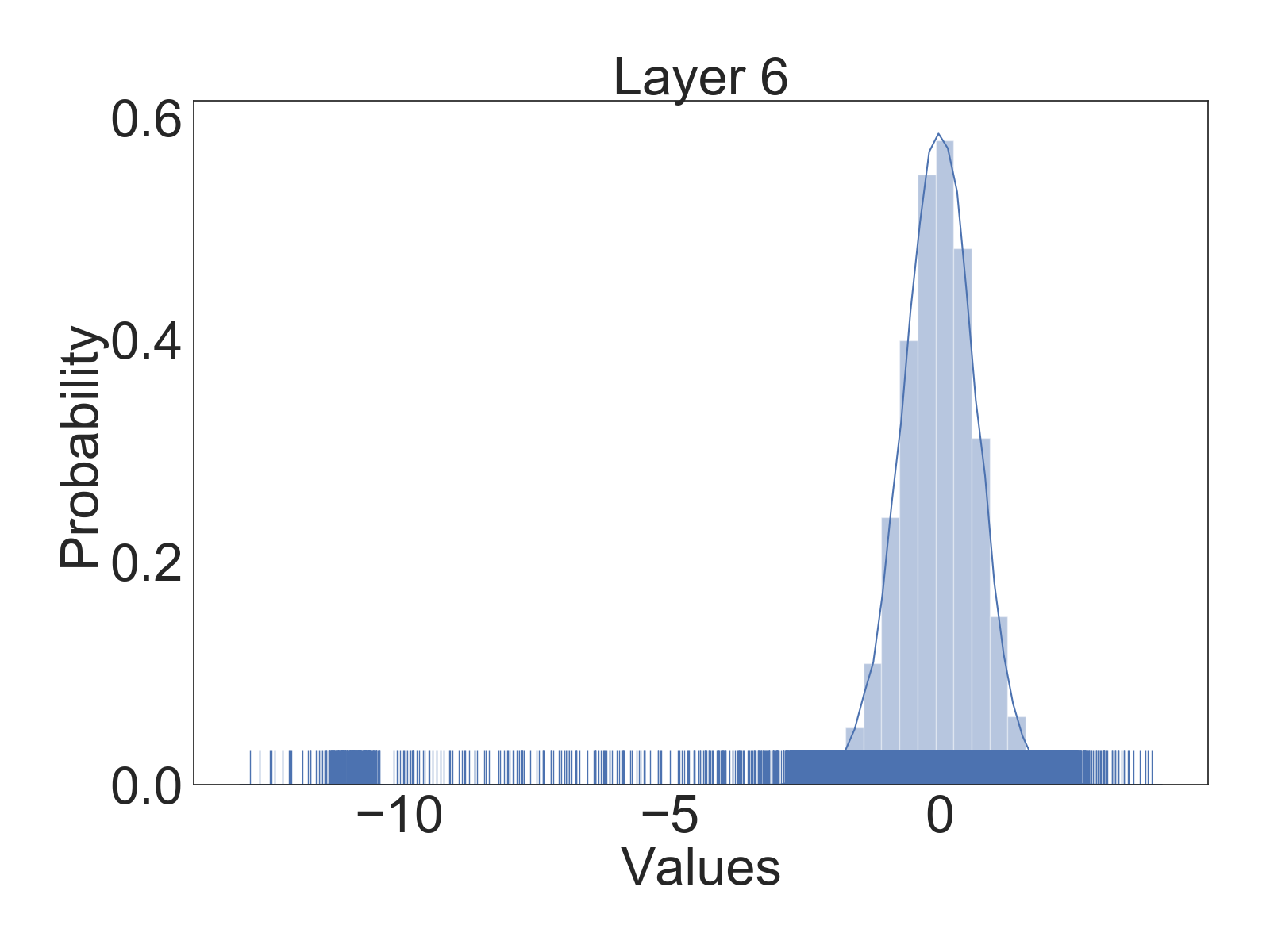}
		\vspace{-0.35in}
	\caption{Distribution of the 1st and 6th Transformer layer's hidden representation of the full-precision BERT trained on SQuAD v1.1.}
	\label{fig:hidn_rep}
	\vspace{-0.1in}
\end{figure}

We find that the distribution of hidden representations of the Transformer layers in BERT is skewed towards the negative values (Figure~\ref{fig:hidn_rep}). This bias is more obvious for early layers (Appendix~\ref{apdx:hidn_rep_all}).
Thus we use min-max 8-bit  quantization for activations as it gives finer resolution for non-symmetric distributions.
Empirically, we also find that min-max 8-bit quantization outperforms symmetric quantization (Details are in Section~\ref{sec:ablation}). 

Specifically, for one element $x$ in the activation $\x$,
denote $x_{max} = \text{max}(\x)$ and $x_{min} = \text{min}(\x)$,
the min-max 8-bit quantization function is 
\[
Q_a(x) = \text{round}((x - x_{min})/s)\times s + x_{min},
\]
where $
s = (x_{max} - x_{min})/255,
$ is the scaling parameter.
We use the straight-through estimator in \citep{courbariaux2016binarized} to back propagate the gradients through the quantized activations.

\subsection{Distillation-aware Ternarization}
\label{sec:distill}
The quantized BERT uses low bits to represent the model parameters and activations. Therefore it results in relatively low capacity and worse performance compared with the full-precision counterpart. To alleviate  this problem, we incorporate the technique of knowledge distillation to improve performance of the quantized BERT. In this teacher-student knowledge distillation framework, the quantized BERT acts as the student model, and learns to recover the behaviours of the full-precision teacher model over the Transformer layers and prediction layer.

Specifically, inspired by \citet{jiao2019tinybert},
the distillation objective for the Transformer layers $\mathcal{L}_{trm}$ consists of two parts. The first part is the distillation loss 
which distills knowledge in the embedding layer and the outputs of all Transformer layers of the full-precision teacher model to the quantized student model,
by the mean squared error (MSE) loss:
$\sum_{l=1}^{L+1} \text{MSE}(\H_l^S,\H_l^T).
$
The second part is the distillation loss
that distills knowledge from
the teacher model's attention scores from all heads $\A_l^T$ in each Transformer layer to the student model's attention scores $\A_l^S$ as $\sum_{l=1}^{L}\text{MSE}(\A_l^S,\A_l^T)$.
Thus the distillation for the Transformer layers $\mathcal{L}_{trm}$ is formulated as:
\begin{equation*}
\mathcal{L}_{trm} \!\! = \!\!\sum_{l=1}^{L+1} \text{MSE}(\H_l^S,\H_l^T)  \!+\! \sum_{l=1}^{L}\text{MSE}(\A_l^S,\A_l^T).
\end{equation*}

Besides the Transformer layers, we also 
 distill knowledge in the prediction layer which makes the student model's logits $\P^S$  learn to fit $\P^T$ from the  teacher model by the soft cross-entropy (SCE) loss:
\begin{equation*}
\mathcal{L}_{pred} = \text{SCE}(\P^S, \P^T).
\end{equation*}
 The overall objective of knowledge distillation in the training process of TernaryBERT is thus
\begin{equation}
\mathcal{L} = \mathcal{L}_{trm} + \mathcal{L}_{pred}.\label{eq:distill}
\end{equation}

We use the full-precision BERT  fine-tuned on the downstream task
to initialize our quantized model, and
 the data augmentation method in \citep{jiao2019tinybert} to boost the performance.
The whole procedure, which will be called Distillation-aware ternarization,  is shown in Algorithm~\ref{alg:main}. 
\begin{algorithm}[h!]
	\caption{Distillation-aware ternarization.}
	\label{alg:main}
	{\bf initialize:} A fixed teacher model and a trainable student model  using a  fine-tuned BERT model. \\
	{\bf input:} (Augmented) training data set. \\
	{\bf output:} TernaryBERT $\hw$.
	
	\begin{algorithmic}[1]
		\FOR {$t = 1, ..., T_{train} $}
		\STATE Get next mini-batch of data;
		\STATE Ternarize $\w^t$ in student model  to
		$\hw^t$;
		\STATE Compute distillation loss $\mathcal{L}$ in ~(\ref{eq:distill});
		\STATE  Backward propagation of the student model and compute 
		the gradients $\frac{\partial \mathcal{L}}{\partial \hw^t}$;
		\STATE $\w^{t+1} = \text{UpdateParameter}(\w^t, \frac{\partial \mathcal{L}}{\partial \hw^t}, \eta^t)$;
		\STATE $\eta^{t+1} = \text{UpdateLearningRate}(\eta^t, t)$.
		\ENDFOR
	\end{algorithmic}
\end{algorithm}

\section{Experiments}

\newcommand{\tabincell}[2]{\begin{tabular}{@{}#1@{}}#2\end{tabular}}
\begin{table*}[htbp]
	\caption{Development set results of quantized BERT and TinyBERT on the GLUE benchmark. We abbreviate the number of bits for weights of Transformer layers, word embedding and activations as ``W-E-A (\#bits)". }
	\label{tbl:main_glue}
	\vspace{-0.1in}
	\centering
	\scalebox{0.7}{
		\begin{tabular}{llcc|cccccccc}
			\hline
			                       &                                            & \tabincell{c}{W-E-A\\(\#bits)} & \tabincell{c}{Size\\(MB)} & \tabincell{c}{MNLI-\\m/mm } &        QQP         &     QNLI      &     SST-2     &     CoLA      &        STS-B       &    MRPC           &      RTE      \\ \hline
			                       & BERT                                       &            32-32-32            &     418 ($\times 1$)      &          84.5/84.9          &     87.5/90.9      &     92.0      &     93.1      &     58.1      &     89.8/89.4      &    90.6/86.5         &     71.1    \\ 
			                       & TinyBERT                                   &            32-32-32            &     258 ($\times 1.6$)    &          84.5/84.5          &     88.0/91.1      &     91.1      &     93.0      &     54.1      &     89.8/89.6      &     91.0/87.3      &     71.8      \\\hline\hline
			\multirow{5}{*}{2-bit} & Q-BERT                                     &             2-8-8              &    43  ($\times 9.7$)     &          76.6/77.0          &         -          &       -       &     84.6      &       -       &         -          &         -          &       -       \\
			                       & Q2BERT                                     &             2-8-8              &    43   ($\times 9.7$)    &          47.2/47.3          &     67.0/75.9      &     61.3      &     80.6      &       0       &     4.4/4.7        &      81.2/68.4       &     52.7      \\ \cline{2-12}
			                       & $\text{TernaryBERT}_\text{TWN}$ (ours)     &             2-2-8              &    28 ($\times 14.9$)     &          83.3/83.3          &     86.7/90.1      &     91.1      &     92.8      & 55.7          &     87.9/87.7      &     91.2/87.5      &     72.9\\
			                       & $\text{TernaryBERT}_\text{LAT}$ (ours)     &             2-2-8              &    28  ($\times 14.9$)    &          83.5/83.4          &     86.6/90.1      &     91.5      &     92.5      &     54.3      &     87.9/87.6      &     91.1/87.0      &     72.2      \\
			                       & $\text{TernaryTinyBERT}_\text{TWN}$ (ours) &             2-2-8              &    18  ($\times 23.2$)    &          83.4/83.8          &     87.2/90.5      &     89.9      &     93.0      &     53.0      &     86.9/86.5      &     91.5/88.0      &     71.8      \\ \hline\hline
			\multirow{4}{*}{8-bit} & Q-BERT                                     &             8-8-8              &    106 ($\times 3.9$)     &          83.9/83.8          &         -          &       -       &     92.9      &       -       &         -          &         -          &       -       \\
			                       & Q8BERT                                     &             8-8-8              &    106  ($\times 3.9$)    &             -/-             &     88.0/-         &     90.6      &     92.2      &     58.5      &       89.0/-       &     89.6/-         &     68.8      \\ \cline{2-12}
			                       & 8-bit BERT (ours)                          &             8-8-8              &    106 ($\times 3.9$)     &          84.2/84.7          &     87.1/90.5      &     91.8      &     93.7      & 60.6          &   89.7/89.3          &     90.8/87.3      &     71.8    \\
			                       & 8-bit TinyBERT (ours)                      &             8-8-8              &    65  ($\times 6.4$)     &          84.4/84.6          &     87.9/91.0      &     91.0      &     93.3      &     54.7      &     90.0/89.4      &     91.2/87.5      & 72.2\\ \hline
		\end{tabular}
	}  
\end{table*}

\begin{table*}[htbp]
	\caption{Test set results of the proposed quantized BERT and TinyBERT on the GLUE benchmark.}
	\centering
	\label{tbl:glue}
	\vspace{-0.1in}
	\scalebox{0.73}{
		\begin{tabular}{lcc|ccccccccc}
			\hline
			                      & \tabincell{c}{W-E-A\\(\#bits)} & \tabincell{c}{Size\\(MB)} & \tabincell{c}{MNLI\\(-m/mm)} &    QQP    & QNLI & SST-2 & CoLA &   STS-B   &   MRPC    & RTE  &     score   \\ \hline
			BERT                  &            32-32-32            &     418 ($\times 1$)      &          84.3/83.4           & 71.8/89.6 & 90.5 & 93.4  & 52.0 & 86.7/85.2 & 87.6/82.6 & 69.7 &     78.2   \\ \hline\hline
			$\text{TernaryBERT}_\text{TWN}$&             2-2-32             &   28   ($\times 14.9$)    &          83.1/82.5           & 71.0/88.6 & 90.2 & 93.4  & 50.1 & 84.7/83.1 & 86.9/81.7 & 68.9 & 77.3 \\
		$\text{TernaryBERT}_\text{TWN}$  &             2-2-8              &    28  ($\times 14.9$)    &          83.0/82.2           & 70.4/88.4 & 90.0 & 92.9  & 47.8 & 84.3/82.7 & 87.5/82.6 & 68.4 & 76.9\\
		$\text{TernaryTinyBERT}_\text{TWN}$&             2-2-8              &   18   ($\times 23.2$)    &          83.8/82.7           & 71.0/88.8 & 89.2 & 92.8  & 48.1 & 81.9/80.3 & 86.9/82.2 & 68.6 & 76.6  \\ \hline\hline
			8-bit BERT    &             8-8-8              &    106 ($\times 3.9$)     &          84.2/83.5           & 71.6/89.3 & 90.5 & 93.1  & 51.6 & 86.3/85.0 & 87.3/83.1 & 68.9 & 77.9  \\
			8-bit TinyBERT &             8-8-8              &    65  ($\times 6.4$)     &          84.2/83.2           & 71.5/89.0 & 90.4 & 93.0  & 50.7 & 84.8/83.4 & 87.4/82.8 & 69.7 & 77.7 \\ \hline
		\end{tabular}
	}
\end{table*} 

In this section, we evaluate the efficacy of the proposed TernaryBERT on both the GLUE benchmark~\cite{wang2018glue} and SQuAD~\cite{rajpurkar2016squad,rajpurkar2018know}. 
The experimental code is modified from the huggingface transformer library.\footnote{Given the superior performance of Huawei Ascend AI Processor and MindSpore computing framework, we are going to open source the code based on MindSpore (\url{https://www.mindspore.cn/en}) soon.} 
We use both  TWN and LAT
to ternarize the weights. 
We use layer-wise ternarization for weights in Transformer layers while row-wise ternarization for the word embedding,
because empirically finer granularity to word embedding improves performance (Details are in Section~\ref{sec:ablation}). 

We compare our proposed method with Q-BERT~\citep{shen2019q} and Q8BERT~\citep{zafrir2019q8bert} using their reported results.
We also compare with a weight-ternarized BERT baseline Q2BERT
by modifying the min-max 8-bit quantization to min-max ternarization using the released code of Q8BERT.\footnote{\url{https://github.com/NervanaSystems/nlp-architect.git}} 
For more direct comparison, we also evaluate the proposed method  under the same 8-bit quantization settings as Q-BERT and Q8BERT.
When the weights are quantized to 8-bit, we use layer-wise scaling for both the weights in Transformer layers and the word embedding as 8-bit quantization already has high resolution.

\subsection{GLUE benchmark}
\label{expt:glue}
\paragraph{Setup.}

The GLUE benchmark is a collection of diverse natural language understanding tasks, including textual entailment (RTE), natural language inference (MNLI, QNLI), similarity and paraphrase (MRPC, QQP, STS-B), sentiment analysis (SST-2) and linguistic acceptability (CoLA).
For MNLI, we experiment on both the matched (MNLI-m) and mismatched (MNLI-mm) sections. 
The performance metrics are
Matthews correlation for CoLA, F1/accuracy for MRPC and QQP, Spearman correlation  for 
STS-B, and accuracy for the other tasks.

The batch size is 16 for CoLA and 32 for the other tasks. 
The learning rate starts from $2\times 10^{-5}$ and decays linearly to 0 during 1 epoch if trained with the augmented data while 3 epochs if trained with the original data.
The maximum sequence length is 64 for single-sentence tasks CoLA and SST-2, and 128 for the rest sentence-pair tasks.
The dropout rate for hidden representations and the attention probabilities is 0.1.
Since data augmentation does not improve the performance of STS-B, MNLI, and QQP, it is not used
on these three tasks.

\paragraph{Results on BERT and TinyBERT.}
Table~\ref{tbl:main_glue}  shows the development set results on the GLUE benchmark. 
From Table~\ref{tbl:main_glue}, we find that: 1) For ultra-low 2-bit weight, there is a big gap between the Q-BERT (or Q2BERT) and full-precision BERT due to the dramatic reduction in model capacity. TernaryBERT significantly outperforms Q-BERT and Q2BERT, even with  fewer number of bits for word embedding. Meanwhile, TerneryBERT achieves comparable performance with the full-precision baseline with $14.9 \times $ smaller size. 2) When the number of bits for weight increases to 8, the performance of all quantized models is greatly improved and is even comparable as the full-precision baseline, which indicates that the setting  `8-8-8' is not challenging for BERT. Our proposed method outperforms Q-BERT on both MNLI and SST-2 and outperforms Q8BERT in 7  out of 8 tasks. 3) TWN and LAT achieve similar results on all tasks, showing that both ternarization methods are competitive.

In Table~\ref{tbl:main_glue},
we also apply our proposed quantization method on a
6-layer TinyBERT~\cite{jiao2019tinybert} with hidden size of 768, which is trained using distillation.
As can be seen, the quantized 8-bit TinyBERT and TernaryTinyBERT achieve comparable performance as the full-precision baseline.

Test set results 
are summarized in Table~\ref{tbl:glue}.
The proposed TernaryBERT or TernaryTinyBERT achieves comparable scores as the full-precision baseline. Specially,
the TernaryTinyBERT has only 1.6 point accuracy drop while being 23.2x smaller.

\subsection{SQuAD}
\paragraph{Setup.}
	SQuAD v1.1 is a machine reading comprehension task.  Given a question-passage pair, the task is to extract the answer span from the passage. SQuAD v2.0 is an updated version
	 where the question might be unanswerable.
The performance metrics are
EM (exact match) and F1. 

The learning rate decays from $2\times 10^{-5}$ linearly to 0 during 3 epochs. The batch size is 16, and the maximum sequence length is 384.
The dropout rate for the hidden representations and attention probabilities is 0.1.
Since $\mathcal{L}_{trm}$ is several magnitudes larger than $\mathcal{L}_{pred}$ in this task, we separate the distillation-aware quantization into two stages, i.e.,  first using
$\mathcal{L}_{trm}$ as the objective and then $\mathcal{L}$ in (\ref{eq:distill}).

\paragraph{Results.}
Table~\ref{tbl:main_squad} shows the results on SQuAD v1.1 and  v2.0.
TernaryBERT significantly outperforms Q-BERT and Q2BERT, and is even comparable as the full-precision baseline. 
For this
task,  LAT performs slightly better than TWN.

 \begin{table}[htbp]
 		\vspace{-0.05in}
	\caption{Development set results on SQuAD. }
	\label{tbl:main_squad}
	\centering
	\vspace{-0.1in}
	\scalebox{0.7}{
		\begin{tabular}{lcc|cc}
			\hline
			             & \tabincell{c}{W/E/A\\(\#bits)} & \tabincell{c}{Size\\(MB)} & \tabincell{c}{SQuAD\\v1.1} & \tabincell{c}{SQuAD\\v2.0} \\ \hline
			BERT         &           32-32-32            &            418            &         81.5/88.7          & 74.5/77.7                      \\ \hline
			Q-BERT       &             2-8-8             &            43             &         69.7/79.6          & -                          \\
			Q2BERT       &             2-8-8             &            43             &         -          & 50.1/50.1                         \\
		$\text{TernaryBERT}_\text{TWN}$   &             2-2-8             &            28             &         79.9/87.4          & 73.1/76.4                       \\
		$\text{TernaryBERT}_\text{LAT}$   &             2-2-8             &            28             &         80.1/87.5          & 73.3/76.6                \\
			\hline
		\end{tabular}
	}
\vspace{-0.1in}
\end{table}

\subsection{Ablation Study}
\label{sec:ablation}
In this section, we perform ablation study on quantization, knowledge distillation, initialization, and data augmentation.
\paragraph{Weight Ternarization Granularity.}
We evaluate the effects of different granularities (i.e., row-wise and layer-wise ternarization in Section~\ref{sec:ternary}) of TWN on the word embedding and weights in Transformer layers.
The results are summarized in Table~\ref{tbl:sensitivity}. 
There is a gain of using row-wise ternarization over layer-wise ternarization  for word embedding.
We speculate this is because word embedding requires finer granularity as each word contains different semantic information.
For weights in the Transformer layers, layer-wise ternarization performs slightly better than row-wise quantization.
We speculate this is due to high redundancy in the weight matrices, and using one scaling parameter per matrix already recovers most of the representation power of Transformer layers. Appendix~\ref{apdx:attention_pattern} shows that the attention maps of TernaryBERT (with layer-wise ternarization for weights in Transformer layers) resemble the full-precision BERT. Thus empirically, we use row-wise ternarization for word embedding and layer-wise ternarization for weights in the Transformer layers.


\begin{table}[htbp]
			\vspace{-0.05in}
	\centering
	\caption{Development set results of $\text{TernaryBERT}_\text{TWN}$
		 with different ternarization granularities on weights in Transformer layers and word embedding. }
	\label{tbl:sensitivity}
		\vspace{-0.1in}
	\scalebox{0.8}{
	\begin{tabular}{cccc}
		\hline
		Embedding  & Weights    & MNLI-m &      MNLI-mm      \\ \hline
		layer-wise & layer-wise &  83.0&83.0       \\
		layer-wise & row-wise   &  82.9&82.9    \\
		row-wise   & layer-wise &  \textbf{83.3}  & \textbf{83.3} \\
		row-wise   & row-wise   &  83.2&82.9       \\ \hline
	\end{tabular}
}
\vspace{-0.1in}
\end{table}

\paragraph{Activation Quantization.}

For activations, we experiment on both symmetric and min-max 8-bit quantization with SQuAD v1.1 in Table~\ref{tbl:8bit_quantization}. 
The weights are ternarized using TWN.
As can be seen, the performance of min-max quantization outperforms the symmetric quantization.
As discussed in Section~\ref{sec:ternary}, this may because of the non-symmetric distributions of the hidden representation.
\begin{table}[htbp]
		\vspace{-0.02in}
	\centering
	\caption{Comparison of  symmetric 8-bit  and min-max 8-bit activation quantization methods on SQuAD v1.1. }
	\vspace{-0.1in}
	\label{tbl:8bit_quantization}
	\scalebox{0.8}{
		\begin{tabular}{ccl|cc}
			\hline
			W(\#bit) & E(\#bit) & A(\#bit)       & EM    & F1     \\ \hline
			2     & 2     & 8 (sym)  & 79.0   & 86.9   \\
			2     & 2     & 8 (min-max)      & 79.9   & 87.4  \\ \hline
		\end{tabular}
	}
\vspace{-0.1in}
\end{table}
%

\paragraph{Knowledge Distillation.}
In Table~\ref{tbl:effects_kd}, we investigate the effect of  distillation loss over  Transformer layers (abbreviated as ``Trm") and final output logits (abbreviated as ``logits") in the training of $\text{TernaryBERT}_\text{TWN}$.
As can be seen,
without  distillation over the Transformer layers, the 
performance drops by  3\% or more on CoLA and RTE, and also slightly on MNLI.
The accuracy of all tasks further decreases if distillation  logits is also not used. In particular, the accuracy for CoLA, RTE and SQuAD v1.1 drops by over
5\% by removing the distillation compared to the counterpart.

\begin{table}[htbp]
			\vspace{-0.02in}
	\centering
	\caption{Effects of knowledge distillation on the Transformer layers and logits on $\text{TernaryBERT}_\text{TWN}$. ``-Trm-logits" means we use cross-entropy loss w.r.t.  the ground-truth labels as the training objective.}
		\vspace{-0.1in}
	\label{tbl:effects_kd}
	\scalebox{0.75}{
		\begin{tabular}{l|cccc}
			\hline
			                 & MNLI-m/mm & CoLA &   RTE  & SQuADv1.1 \\ \hline
			TernaryBERT      &         83.3/83.3          & 55.7 &  72.9 &         79.9/87.4          \\ \hline
			\quad-Trm        &         82.9/83.3          & 52.7 &  69.0 &         76.6/84.9          \\ \hline
			\quad-Trm-logits &         80.8/81.1          & 45.4 & 56.3 &         74.3/83.2          \\ \hline
		\end{tabular}
	}
	\vspace{-0.1in}
\end{table}

\paragraph{Initialization and Data Augmentation.}
Table~\ref{tbl:effects_da_init} demonstrates the effect of  initialization from a fine-tuned BERT otherwise a pre-trained BERT, and the use
of data augmentation in training TernaryBERT. 
As can be seen, both factors contribute positively to the performance and the improvements are more obvious on CoLA and RTE.

\begin{table}[htbp]
				\vspace{-0.02in}
	\centering
	\caption{Effects of data augmentation and initialization.}
			\vspace{-0.1in}
	\label{tbl:effects_da_init}
	\scalebox{0.8}{
		\begin{tabular}{l|ccccc}
			\hline
			              & CoLA &   MRPC    & RTE   \\ \hline
			TernaryBERT  & 55.7 & 91.2/87.5 & 72.9 \\ \hline
			\quad-Data augmentation      & 50.7 & 91.0/87.5 & 68.2            \\ \hline
			\quad-Initialization    & 46.0 & 91.0/87.2 & 66.4    \\ \hline
		\end{tabular}
	}
			\vspace{-0.1in}
\end{table}		

\subsection{Comparison with Other Methods}
\label{sec:comp_other}
In Figure~\ref{fig:comp} and Table~\ref{tbl:comp_other}, we compare the proposed TernaryBERT with 
(i) Other Quantization Methods: including mixed-precision Q-BERT~\cite{shen2019q}, post-training quantization GOBO~\cite{zadeh2020gobo}, as well as Quant-Noise which uses product quantization~\cite{fan2020training}; and
(ii) Other Compression Methods:
including weight-sharing method ALBERT~\cite{lan2019albert},
pruning method LayerDrop~\cite{fan2019reducing}, distillation methods DistilBERT and TinyBERT~\cite{sanh2019distilbert,jiao2019tinybert}.
The result of DistilBERT is taken from \cite{jiao2019tinybert}.
The results for the other methods are taken from their original paper.
To compare with the other mixed-precision methods which use 3-bit weights,
we also extend the proposed method to allow 3 bits (the corresponding model abbreviated as 3-bit BERT, and 3-bit TinyBERT) by replacing LAT with 3-bit Loss-aware Quantization (LAQ)~\cite{hou2018loss}.
The red markers in Figure~\ref{fig:comp} are our results with settings 1) 2-2-8 TernaryTinyBERT, 2) 3-3-8 3-bit TinyBERT and 3) 3-3-8 3-bit BERT.

\begin{table}[htbp]
	\caption{Comparison between the proposed method and other compression methods on MNLI-m. 
		Note that  Quant-Noise uses Product Quantization (PQ) and does not have specific number of bits for each value.}
	\label{tbl:comp_other}
	\vspace{-0.1in}
	\centering
	\scalebox{0.79}{
		\begin{tabular}{lccc}
			\hline
			Method          & \tabincell{c}{W-E-A\\(\#bits)} & \tabincell{c}{Size\\(MB)} & \tabincell{c}{Accuracy\\(\%)} \\ \hline\hline
			DistilBERT      &            32-32-32            &            250            &             81.6              \\
			TinyBERT-4L        &            32-32-32            &            55             &             82.8              \\
			ALBERT-E64      &            32-32-32            &            38             &             80.8              \\
			ALBERT-E128     &            32-32-32            &            45             &             81.6              \\
			ALBERT-E256     &            32-32-32            &            62             &             81.5              \\
			ALBERT-E768     &            32-32-32            &            120            &             82.0              \\
			LayerDrop-6L    &            32-32-32            &            328            &             82.9              \\
			LayerDrop-3L    &            32-32-32            &            224            &             78.6              \\ \hline
			Quant-Noise    &               PQ               &            38             &             83.6              \\
			Q-BERT          &            2/4-8-8             &            53             &             83.5              \\
			Q-BERT          &            2/3-8-8             &            46             &             81.8              \\
			Q-BERT          &            2-8-8             &          28            &           76.6            \\
			GOBO            &             3-4-32             &            43             &             83.7              \\
			GOBO            &             2-2-32             &            28             &             71.0              \\ \hline
			3-bit BERT (ours)      &             3-3-8             &            41             &             84.2              \\
			3-bit TinyBERT (ours) &             3-3-8             &            25             &             83.7              \\
			$\text{TernaryBERT}$  (ours)    &             2-2-8           &            28             &             83.5              \\
			$\text{TernaryTinyBERT}$  (ours) &             2-2-8             &            18             &             83.4              \\ \hline
		\end{tabular}
	}  
	\vspace{-0.15in}
\end{table}

\paragraph{Other Quantization Methods.}
In mixed precision Q-BERT, weights in Transformer layers with steeper curvature are quantized to 3-bit,  otherwise 2-bit, while word embedding is quantized to
 8-bit.
From Table~\ref{tbl:comp_other}, our proposed method achieves better performance than mixed-precision Q-BERT on MNLI,
using only 2 bits for both the word embedding and the weights 
in the Transformer layers.
Similar observations are also made on SST-2 and SQuAD v1.1 (Appendix~\ref{apdx:qbert_more}).

In GOBO, activations are not quantized. 
From Table~\ref{tbl:comp_other},
even with quantized activations,
 our proposed TernaryBERT outperforms GOBO with 
2-bit weights and is even comparable to GOBO with 3/4 bit mixed-precision weights.

\paragraph{Other Compression Methods.}
From Table~\ref{tbl:comp_other},
compared to other popular BERT compression methods other than quantization,
the proposed method achieves similar or better performance, while being much smaller.

\section{Conclusion}
In this paper, we proposed to use approximation-based and loss-aware ternarization 
to ternarize the weights in the BERT model, with different granularities for word embedding 
and weights in the Transformer layers.
Distillation is also used to reduce the accuracy drop caused by lower capacity due to quantization.
Empirical experiments show that the proposed TernaryBERT outperforms state-of-the-art BERT quantization methods and even
performs comparably as the full-precision BERT.

\bibliography{citation,emnlp2020}
\bibliographystyle{acl_natbib}

\clearpage

\appendix
\section*{APPENDIX}

\section{Distributions of Hidden Representations on SQuAD v1.1}
\label{apdx:hidn_rep_all}
\begin{figure}[htbp]
	\vspace{-0.1in}
	\centering
	\includegraphics[width=0.43\linewidth]{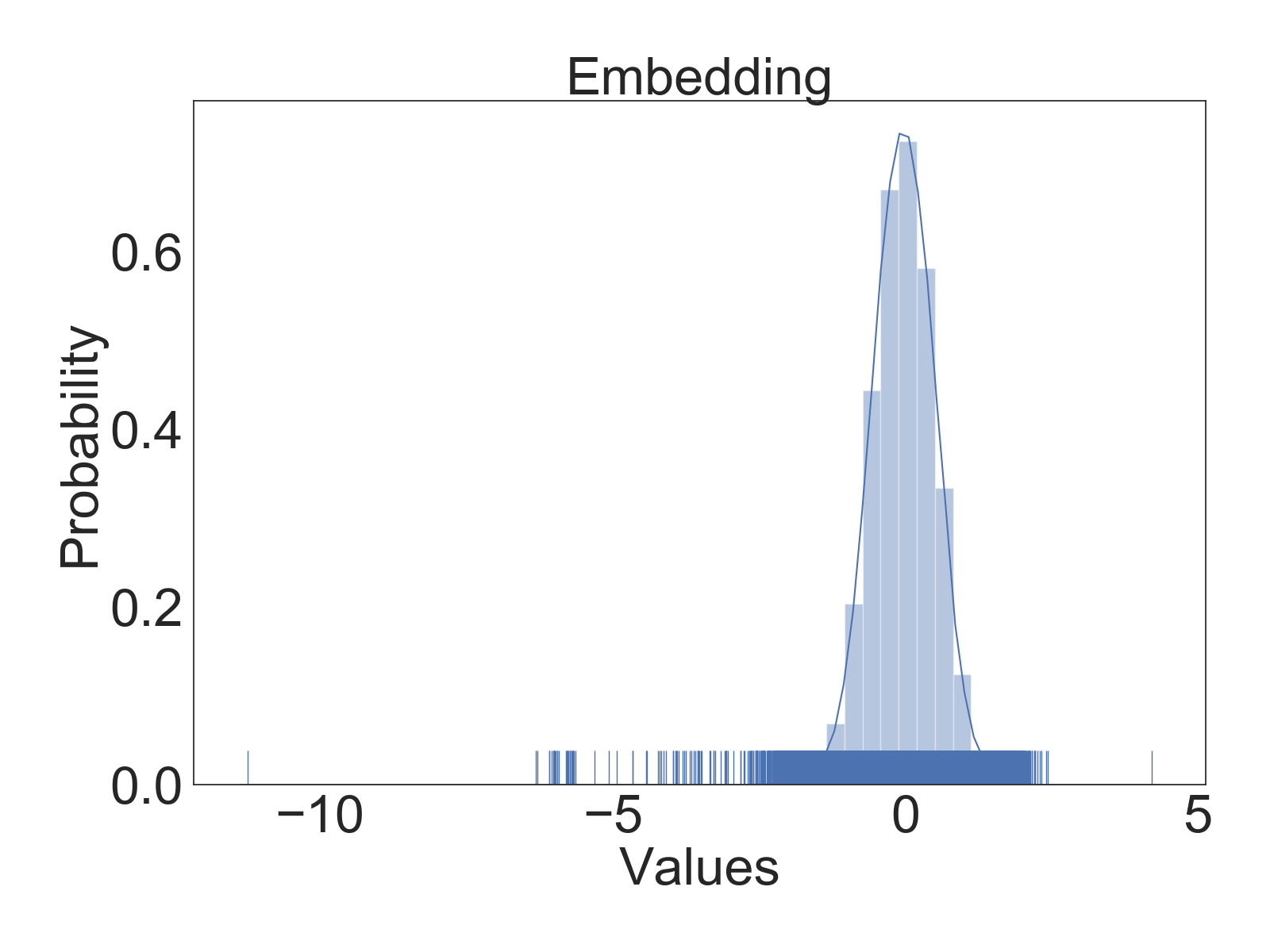}
	\includegraphics[width=0.43\linewidth]{figures/squad_hidden_representations/1.png}\\
	\includegraphics[width=0.43\linewidth]{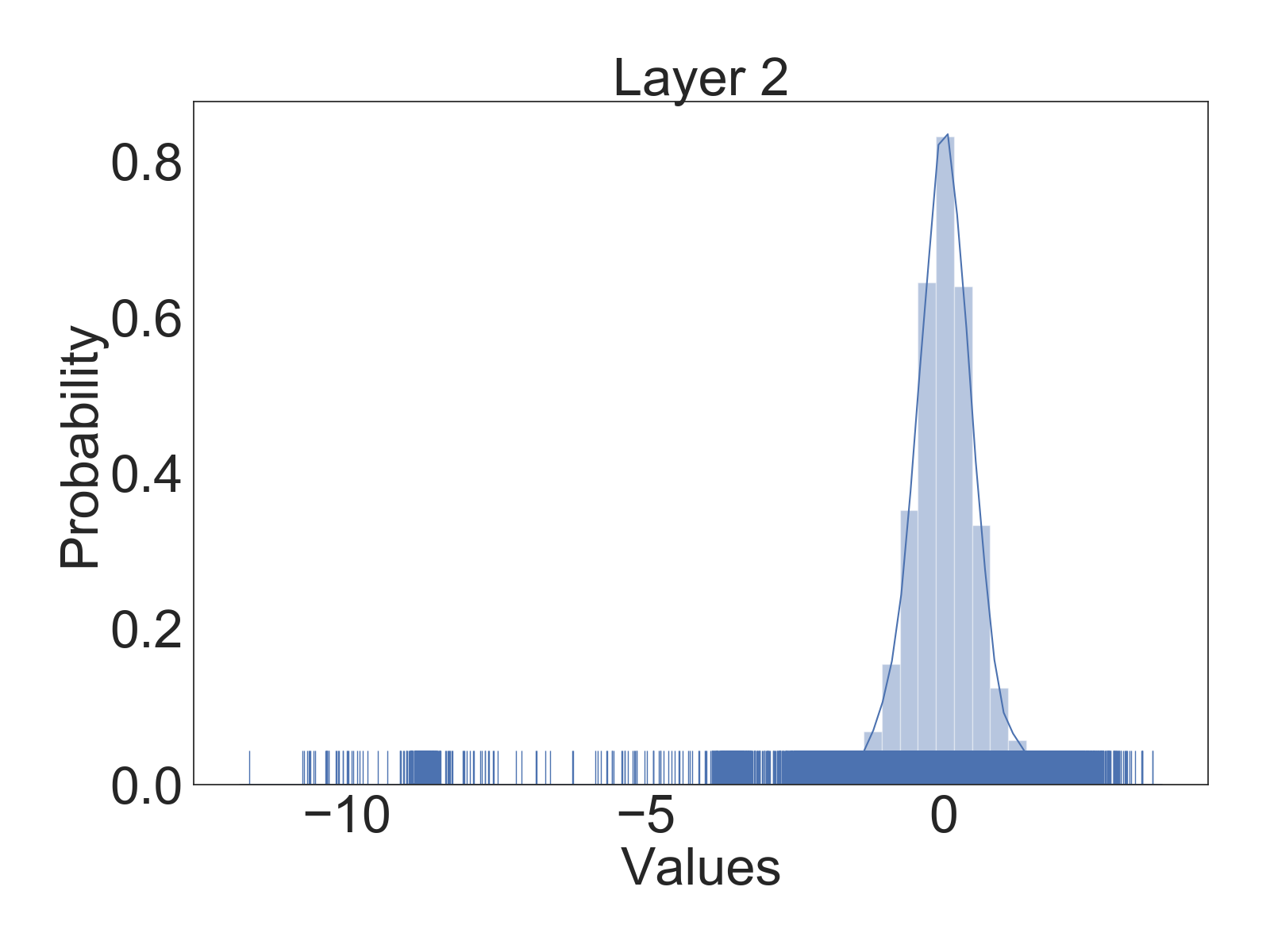}
	\includegraphics[width=0.43\linewidth]{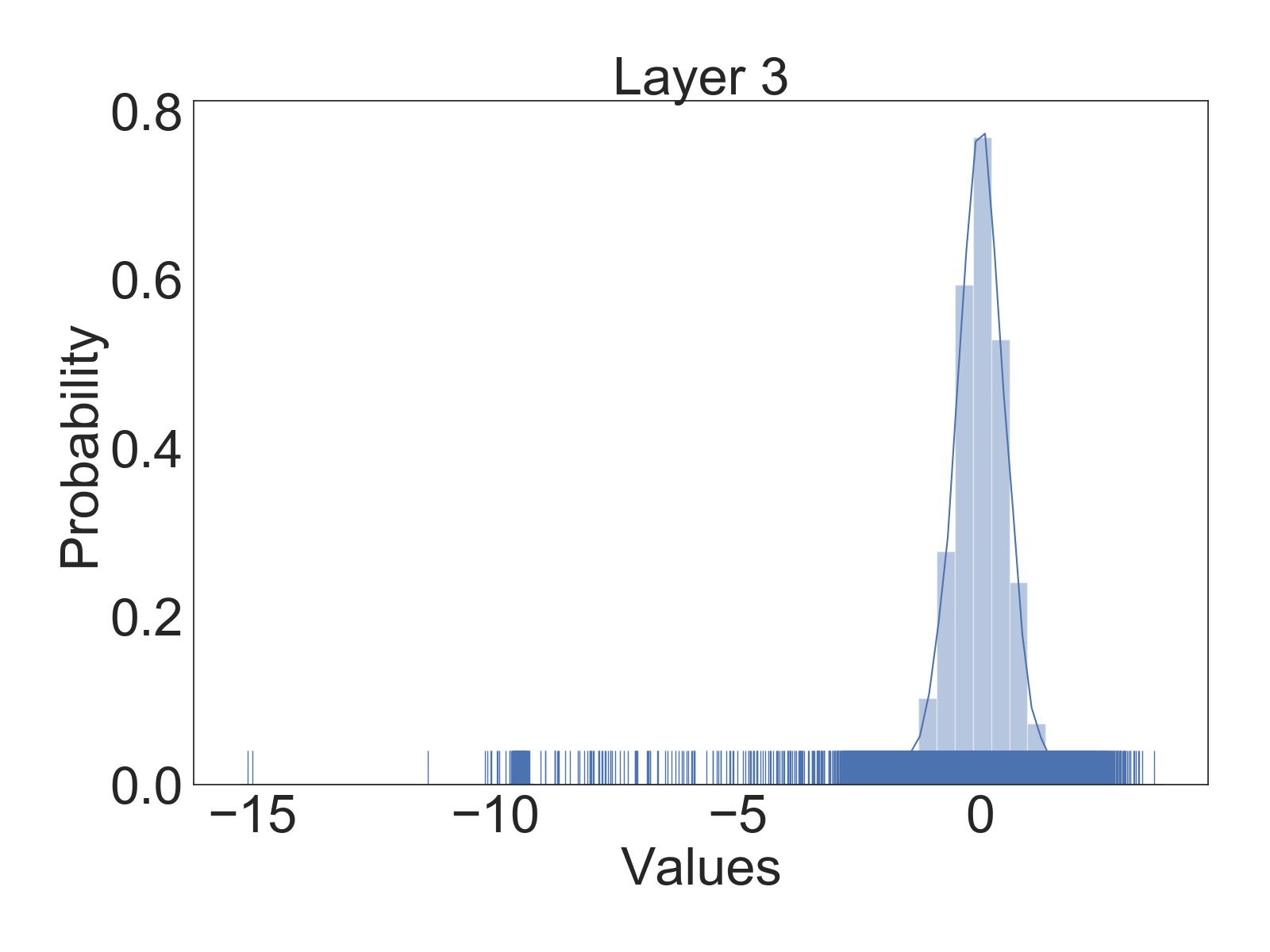}\\
	\includegraphics[width=0.43\linewidth]{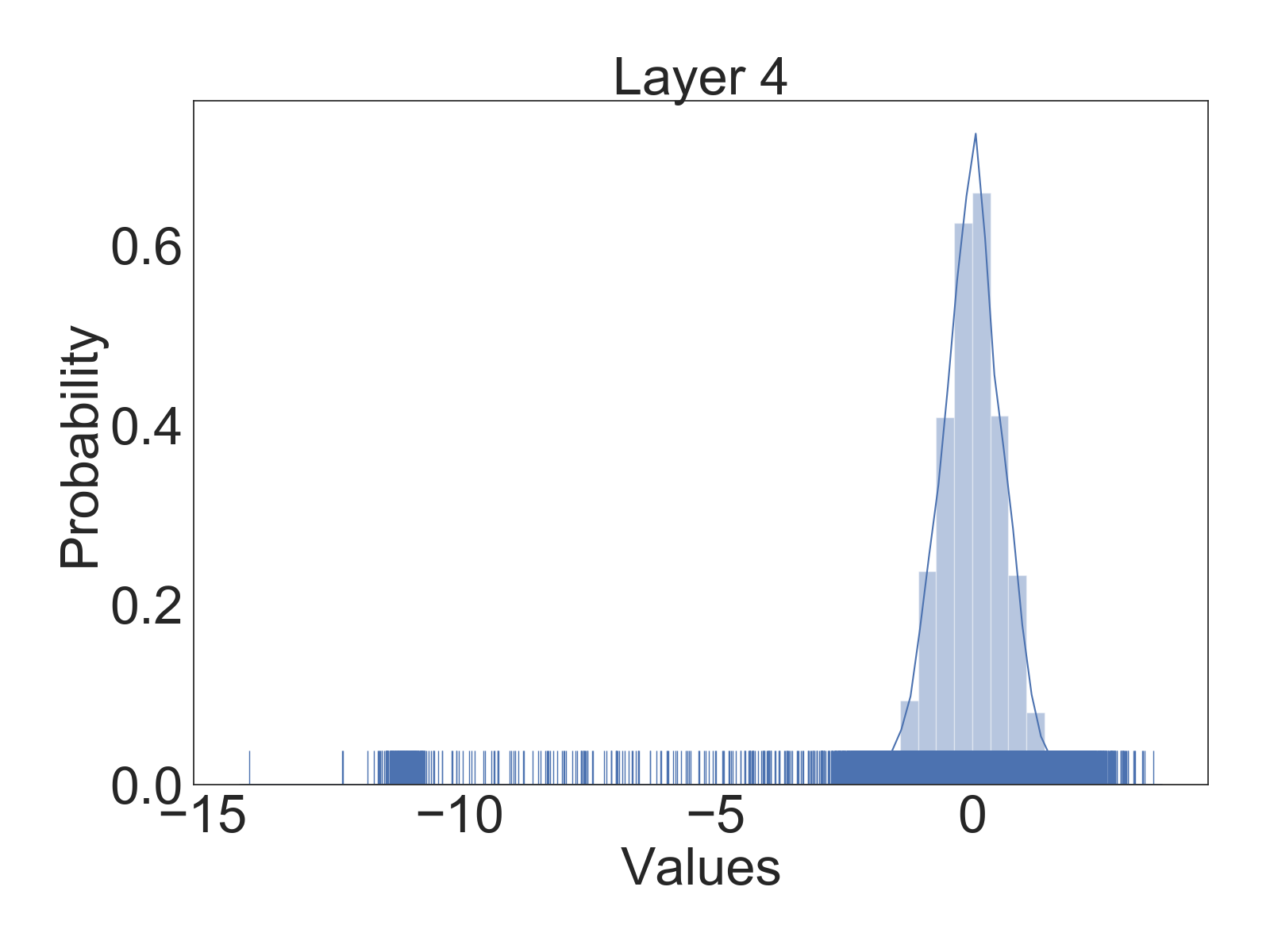}
	\includegraphics[width=0.43\linewidth]{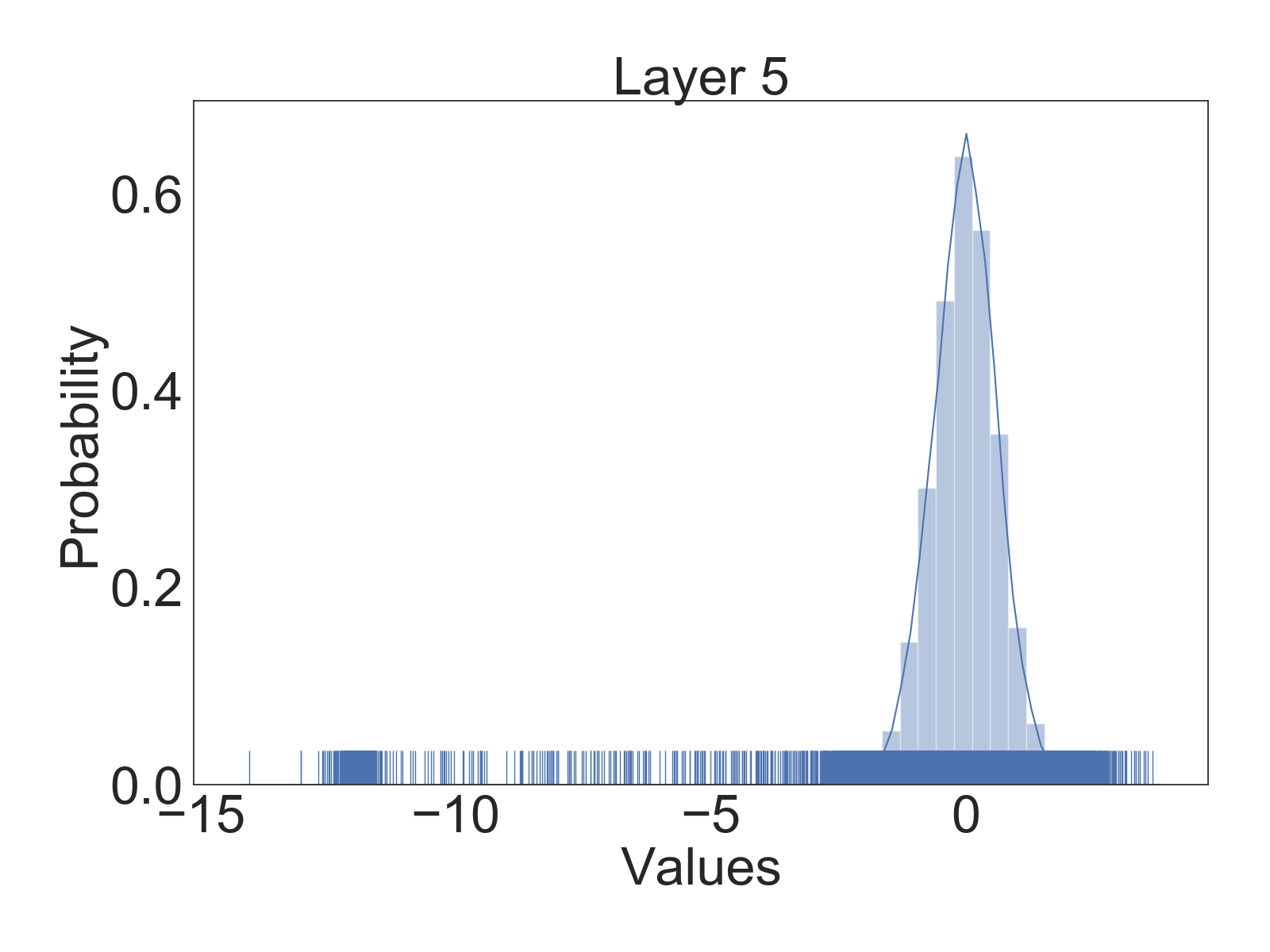}\\
	\includegraphics[width=0.43\linewidth]{figures/squad_hidden_representations/6.png}
	\includegraphics[width=0.43\linewidth]{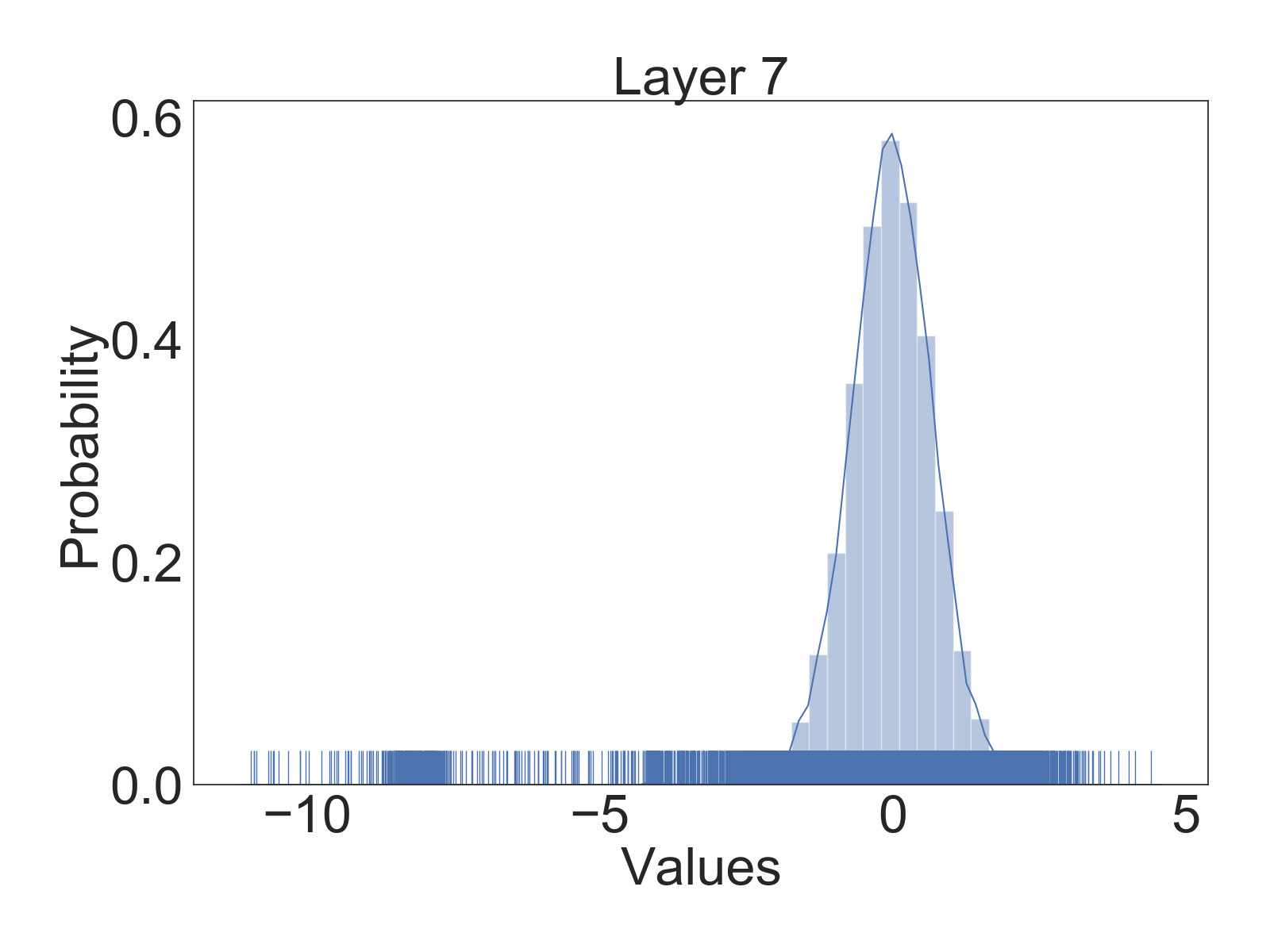}\\
	\includegraphics[width=0.43\linewidth]{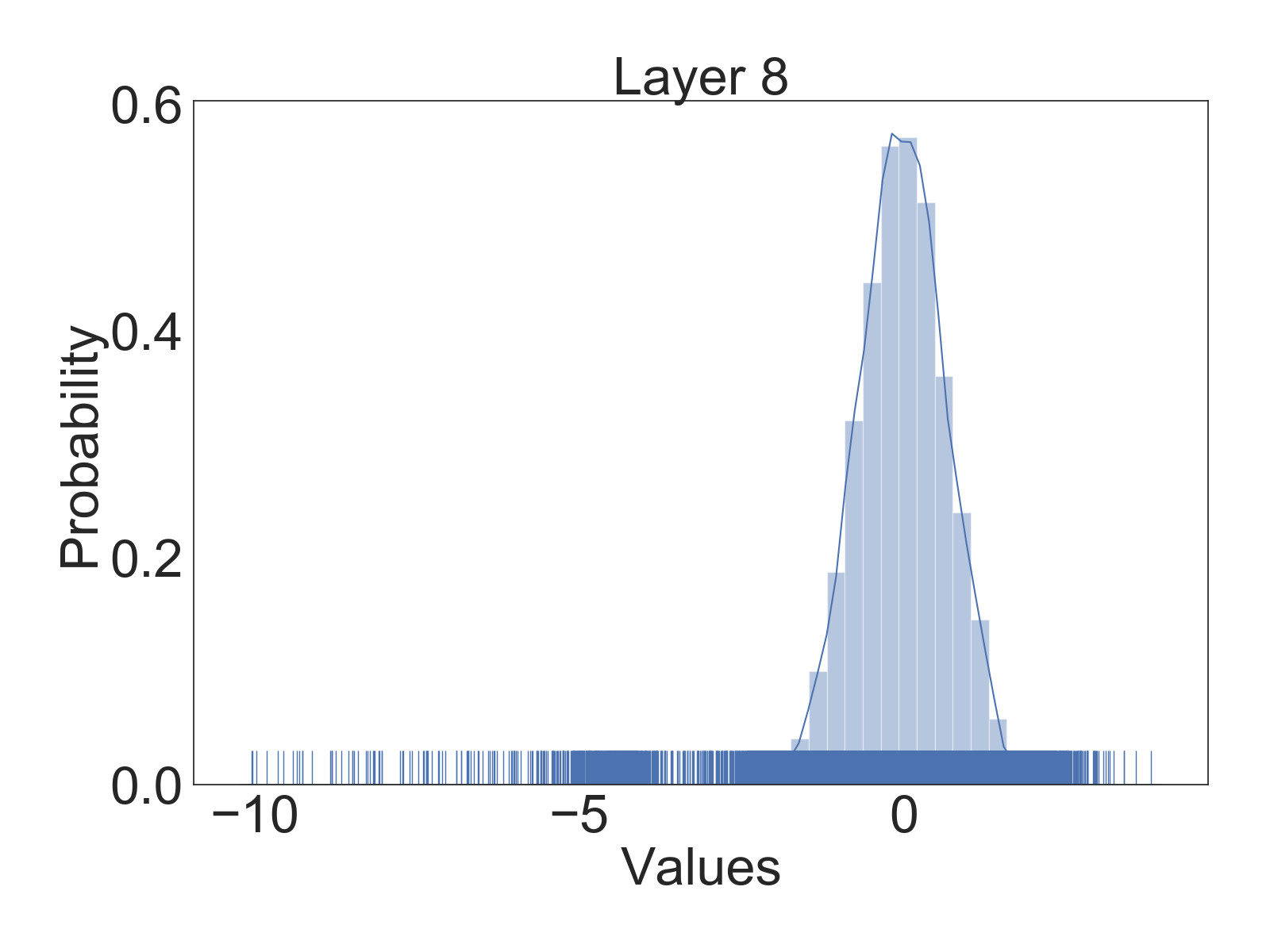}
	\includegraphics[width=0.43\linewidth]{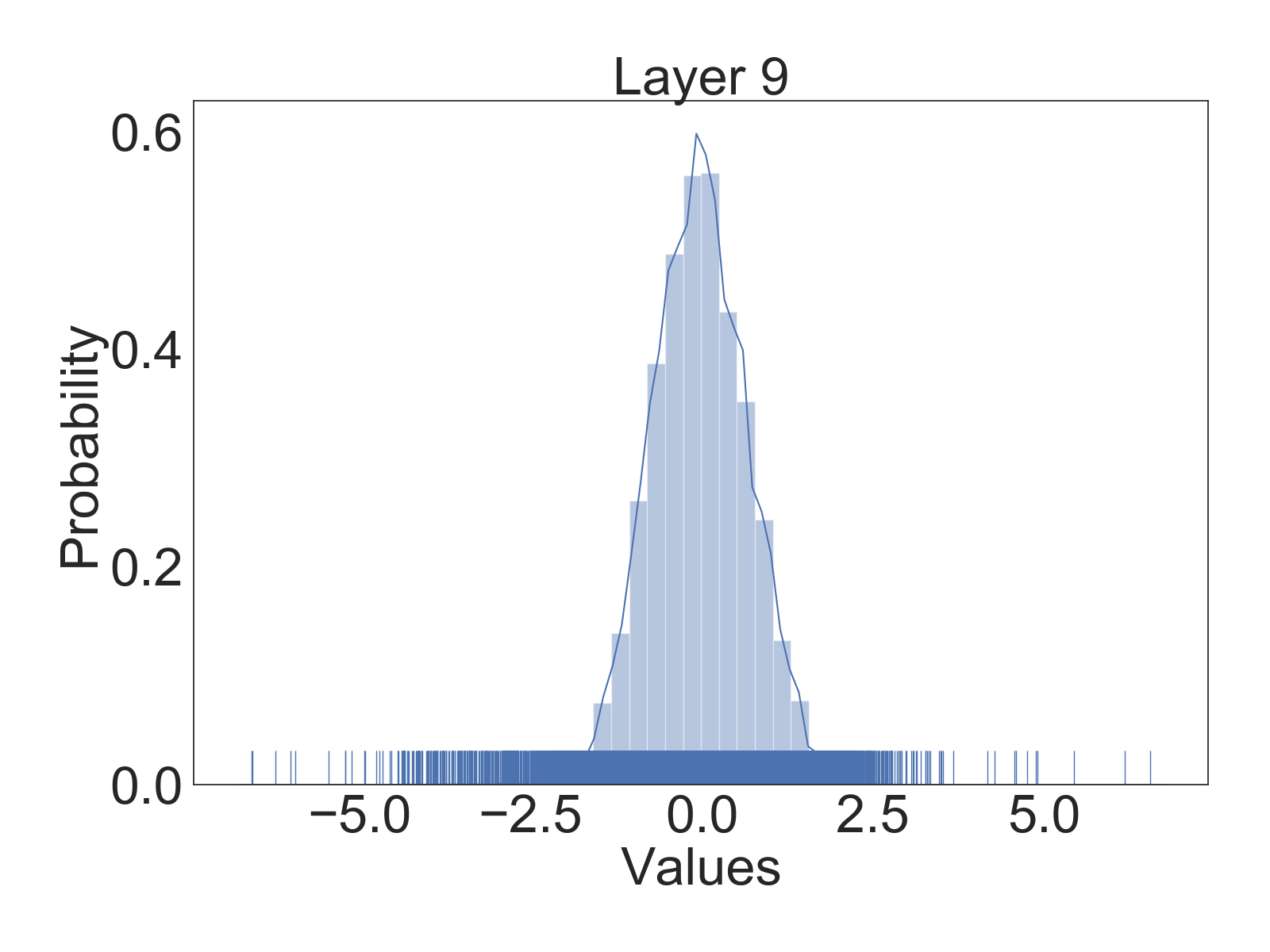}\\
	\includegraphics[width=0.43\linewidth]{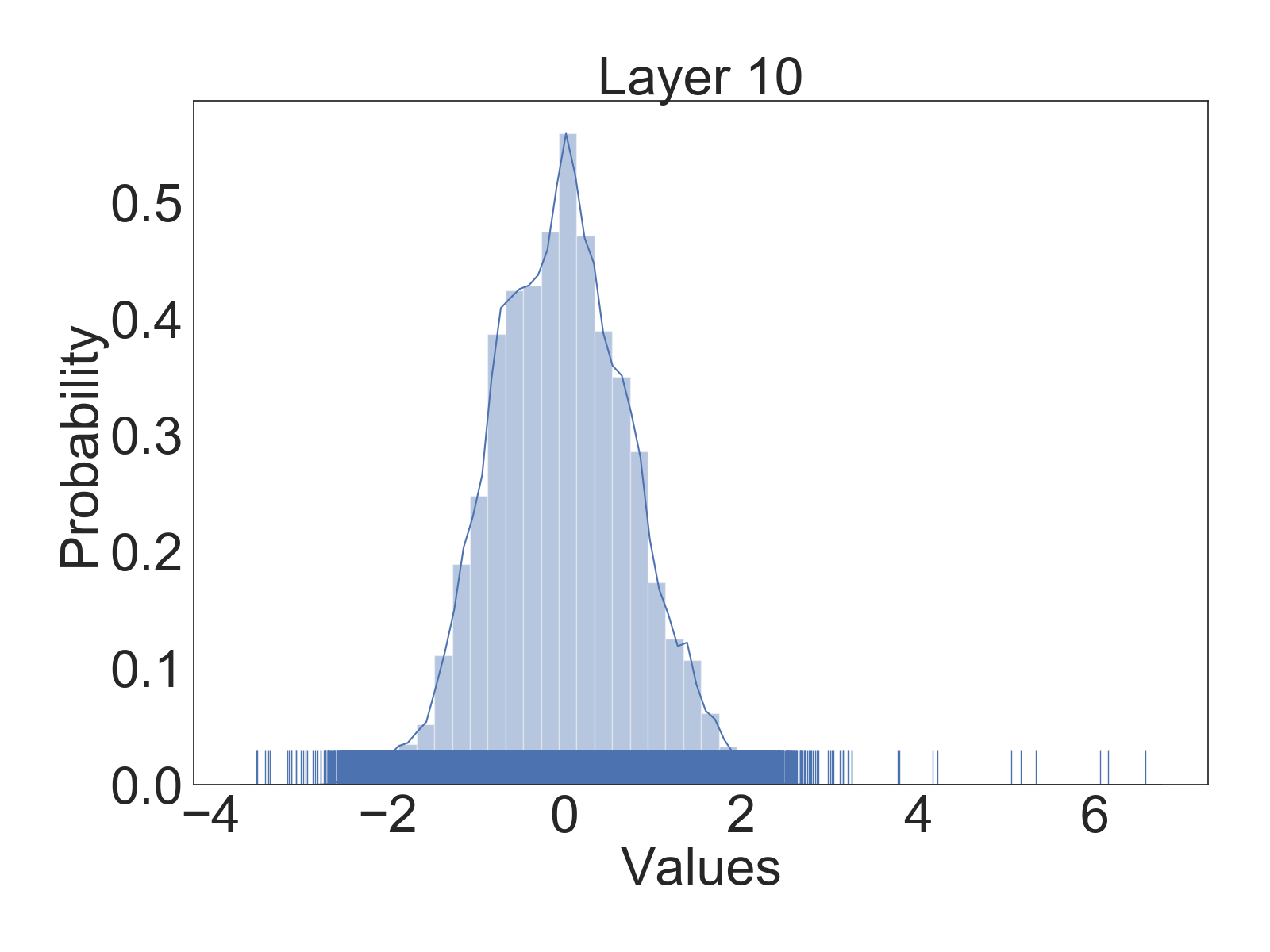}
	\includegraphics[width=0.43\linewidth]{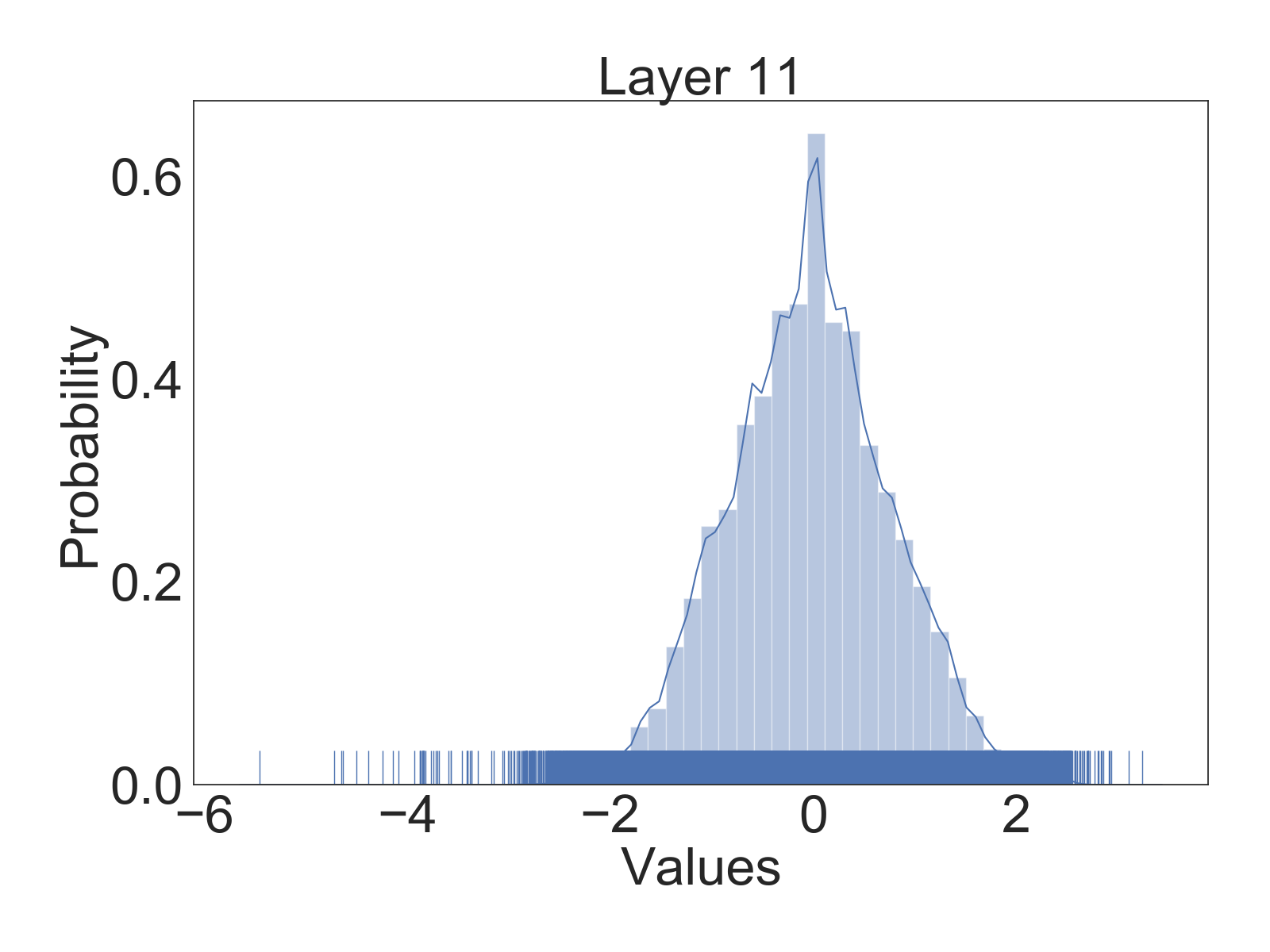}
	\includegraphics[width=0.43\linewidth]{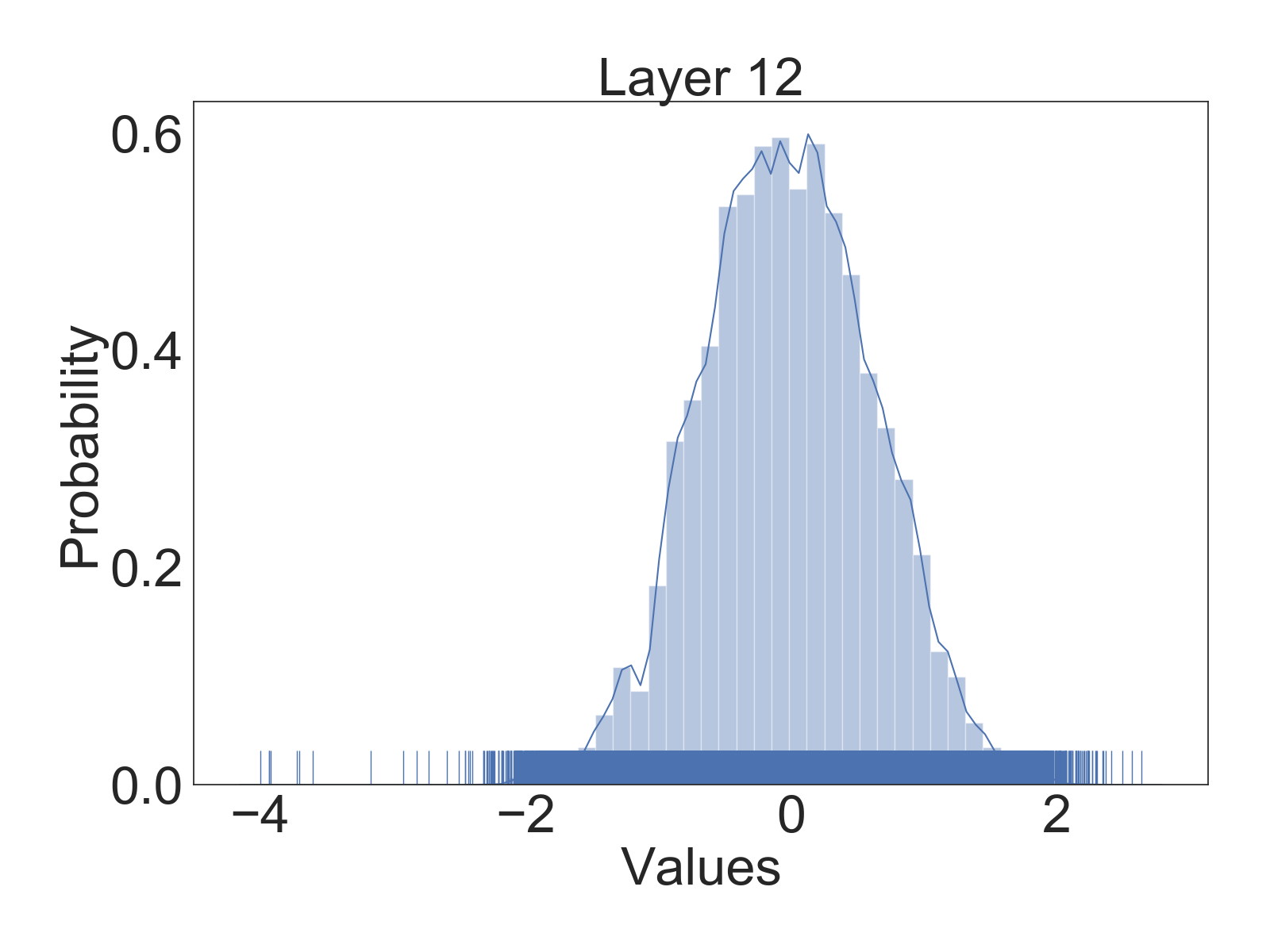}
	\vspace{-0.1in}
	\caption{Distribution of Transformer layer's hidden representation of a full-precision BERT trained on SQuAD v1.1.}
	\label{fig:hidn_rep_all}
\end{figure}

Figure~\ref{fig:hidn_rep_all} shows the distribution of hidden representations from
the embedding layer and  all Transformer layers on SQuAD v1.1. As can be seen, the hidden representations of early layers (e.g. embedding and transformer layers 1-8) are biased towards negative values while those of the rest layers are not.

\section{More Comparison between TernaryBERT and Q-BERT}
\label{apdx:qbert_more}
We compare  with reported results of Q-BERT on SST-2 and SQuAD v1.1 in Table~\ref{tbl:comp_qbert}.
Similar to the observations for MNLI in Section~\ref{sec:comp_other},
our proposed method achieves better performance than mixed-precision Q-BERT on SST-2 and SQuAD v1.1.

\begin{table}[htbp]
	\caption{Comparison between TernaryBERT and mixed-precision Q-BERT. }
	\vspace{-0.1in}
	\label{tbl:comp_qbert}
	\centering
	\scalebox{0.75}{
		\begin{tabular}{lcc|ccc}
			\hline
			& \tabincell{c}{W-E-A\\(\#bits)} & \tabincell{c}{Size\\(MB)} &  SST-2 & \tabincell{c}{SQuAD\\v1.1} \\ \hline\hline
			BERT       &            32-32-32            &            418            &   93.1  &         81.5/88.7          \\ \hline
			Q-BERT     &            2/3-8-8             &            46             &        92.1  &         79.3/87.0          \\
			$\text{TernaryBERT}_\text{TWN}$ &      2-2-8              &            28             &    \textbf{92.8}  &     \textbf{79.9/87.4}          \\ \hline
		\end{tabular}
	}  
	\vspace{-0.15in}
\end{table}

\section{Training Curve on MNLI}
\label{apdx:mnli_training_curve}
\begin{figure}[htbp]	
	\centering
	\vspace{-0.2in}
	\subfloat{
		\includegraphics[width=0.24\textwidth]{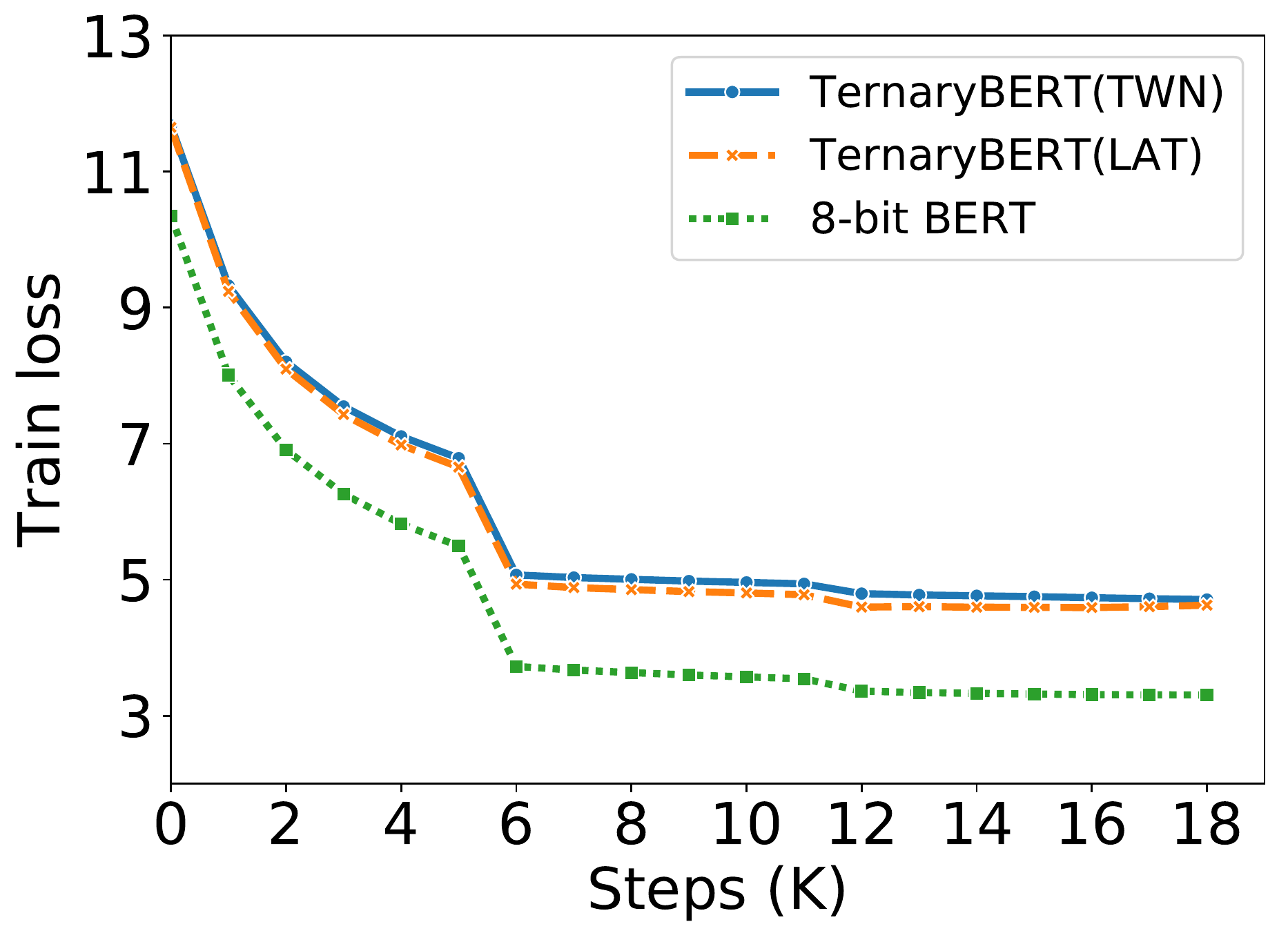}}  
	\subfloat{
		\includegraphics[width=0.24\textwidth]{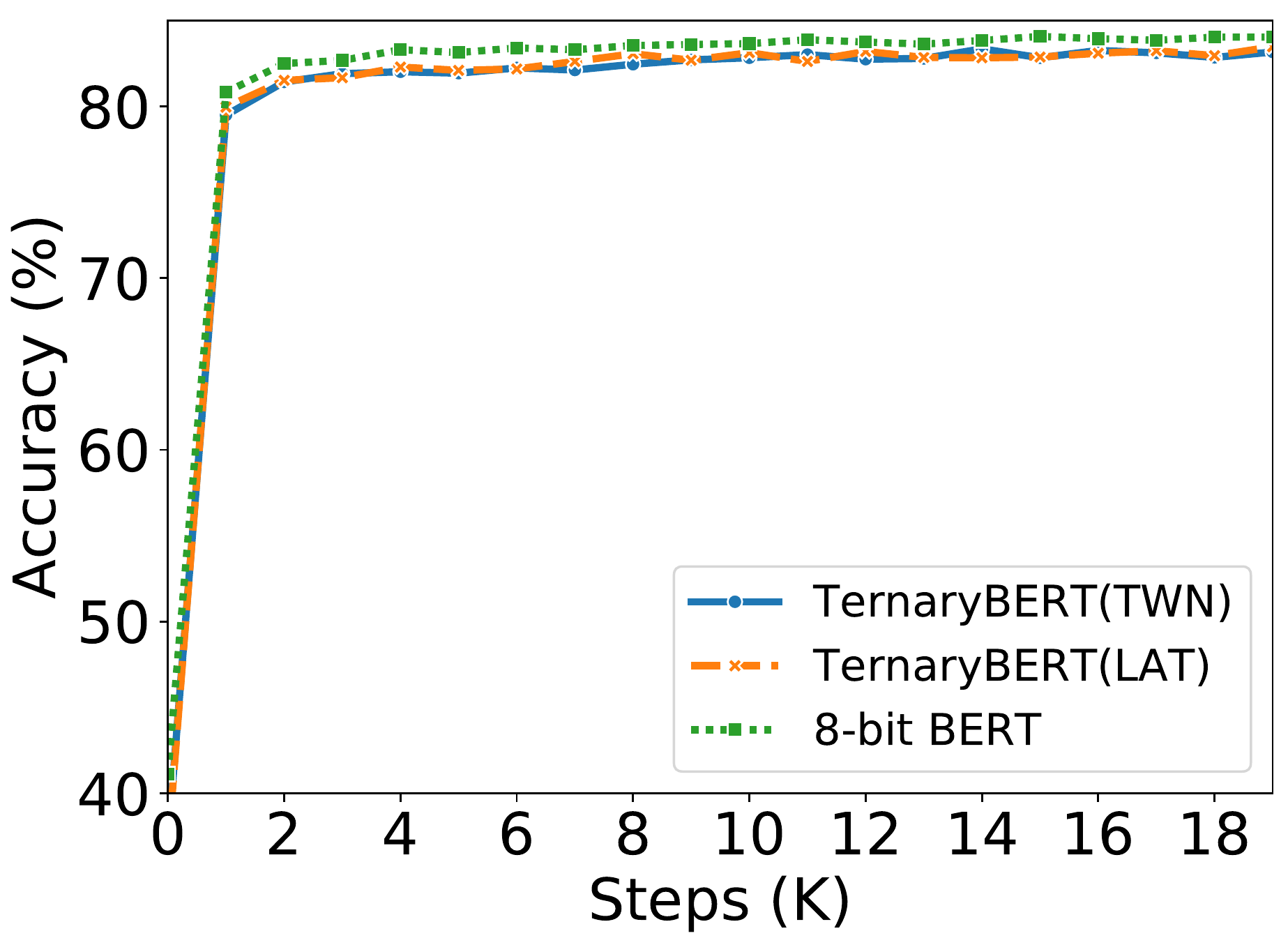}}
	\vspace{-0.1in}
	\caption{Learning curve of TernaryBERT and 8-bit BERT on MNLI-m.}
	\label{fig:mnli_training_curve}
	\vspace{-0.2in}
\end{figure}

Figure~\ref{fig:mnli_training_curve} shows the training loss and validation accuracy of TernaryBERT and 8-bit BERT on MNLI-m. As can be seen,  
8-bit BERT has smaller loss and higher accuracy than TernaryBERT. There is no significant difference between the  learning curve of TernaryBERT using TWN and LAT.

\section{3-bit  BERT and TinyBERT}
In Table~\ref{tbl:3-bit}, we  extend the proposed method to allow 3 bits by replacing LAT with 3-bit Loss-aware Quantization (LAQ).
Compared with $\text{TernaryBERT}_\text{LAT}$, 3-bit BERT  performs lightly better on 7 out of 8 GLUE tasks,
and the accuracy gap with the full-precision baseline is also smaller.
\begin{table*}[htbp]
	\caption{Development set results of 3-bit quantized BERT and TinyBERT on GLUE benchmark.}
	\label{tbl:3-bit}
	\centering
	\scalebox{0.75}{
		\begin{tabular}{lcc|cccccccc}
			\hline
			& \tabincell{c}{W-E-A\\(\#bits)} & \tabincell{c}{Size\\(MB)} & \tabincell{c}{MNLI-\\m/mm } &    QQP      &   QNLI    & SST-2 & CoLA &   MRPC    &   STS-B   &    RTE      \\ \hline
			$\text{TernaryBERT}_\text{LAT}$      &             2-2-8              &            28  ($\times 14.9$)            &          83.5/83.4          & 86.6/90.1 &   91.5    & 92.5  & 54.3 & 91.1/87.0 & 87.9/87.6 &   72.2     \\
			\hline \hline
			3-bit BERT      &             3-3-8              &   41 ($\times 10.2$)      &          \textbf{84.2/84.7}          & 86.9/90.4 &   \textbf{92.0}    & 92.8  & \textbf{54.4} & \textbf{91.3/87.5} & \textbf{88.6/88.3} &   70.8     \\
			3-bit TinyBERT   &             3-3-8              &   25  ($\times 16.7$)      &          83.7/84.0          & \textbf{87.2/90.5}   &   90.7    & \textbf{93.0}  & 53.4 & 91.2/87.3 & 86.1/85.9 &   \textbf{72.6} \\ \hline
		\end{tabular}
	}  
\end{table*}

\section{Attention Pattern of BERT and TernaryBERT}
\label{apdx:attention_pattern}
\begin{figure*}[htbp]
	\centering
	\subfloat[Full-precision BERT.]{
		\includegraphics[width=0.5\textwidth]{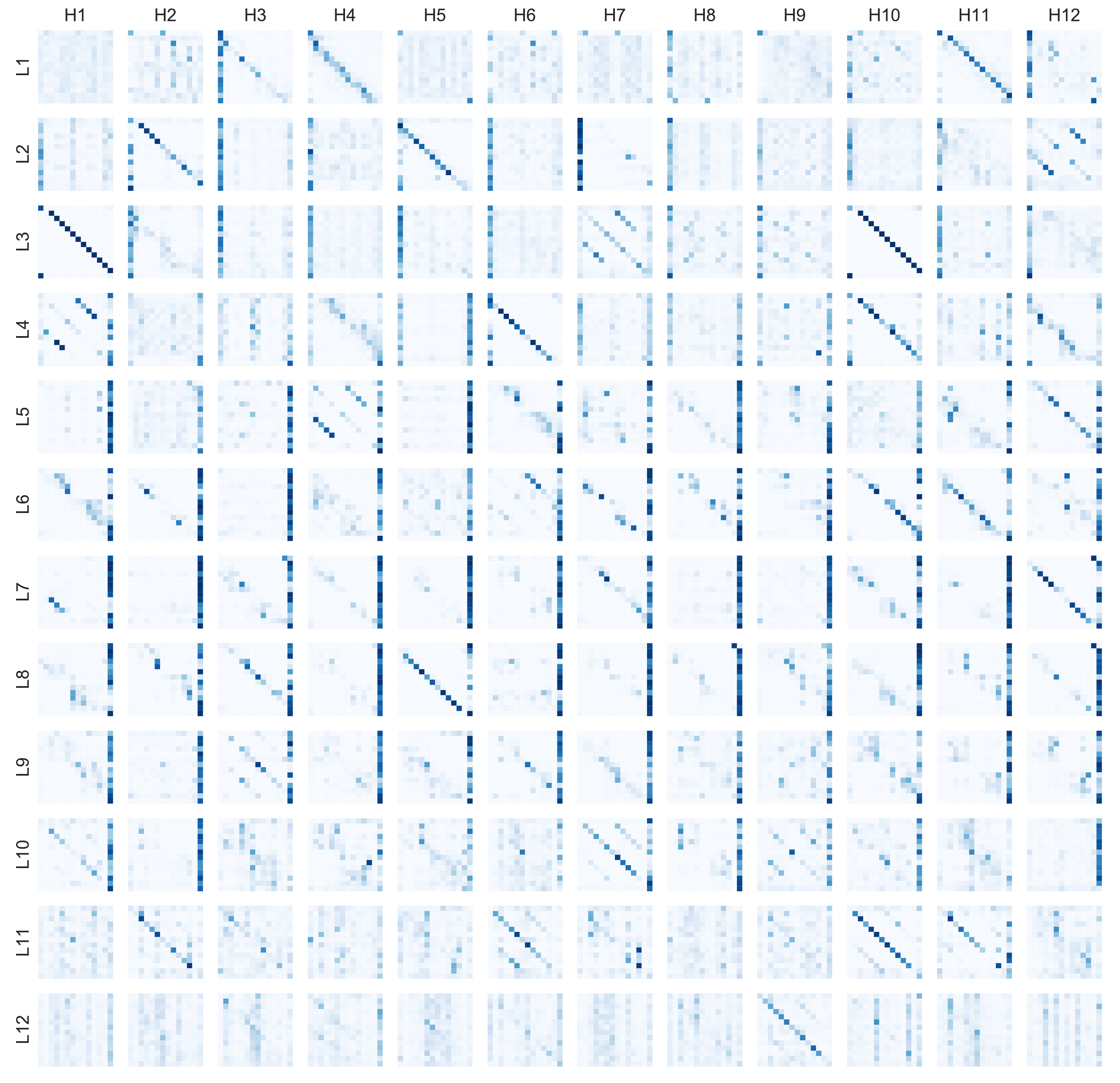}}  
	\subfloat[TernaryBERT.]{
		\includegraphics[width=0.5\textwidth]{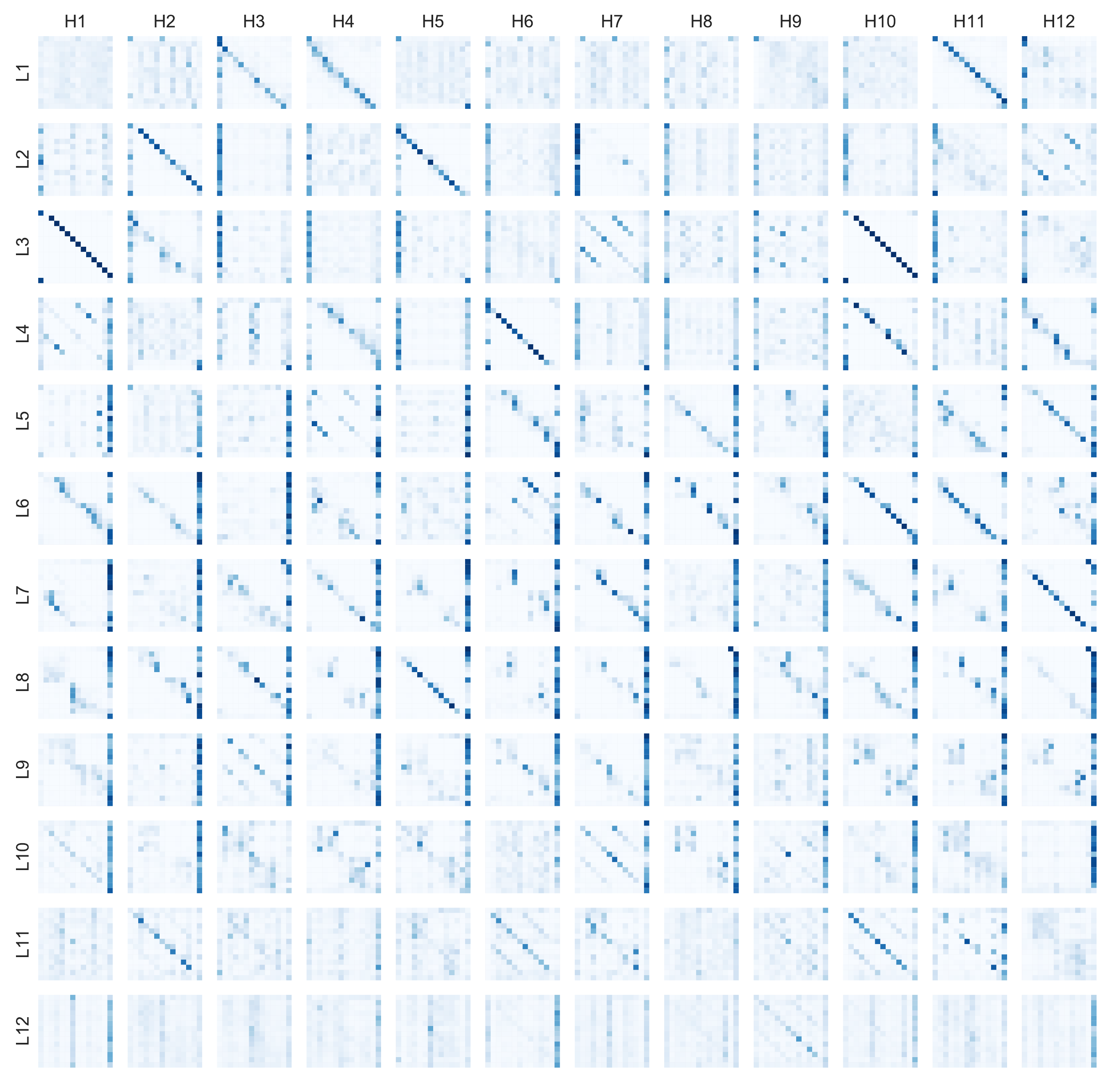}}
	\caption{Attention patterns of full-precision and ternary BERT trained on CoLA. The input sentence is ``The more pictures of him that appear in the news, the more embarrassed John becomes."}
	\label{fig:cola}
	\subfloat[Full-precision BERT.]{
		\includegraphics[width=0.5\textwidth]{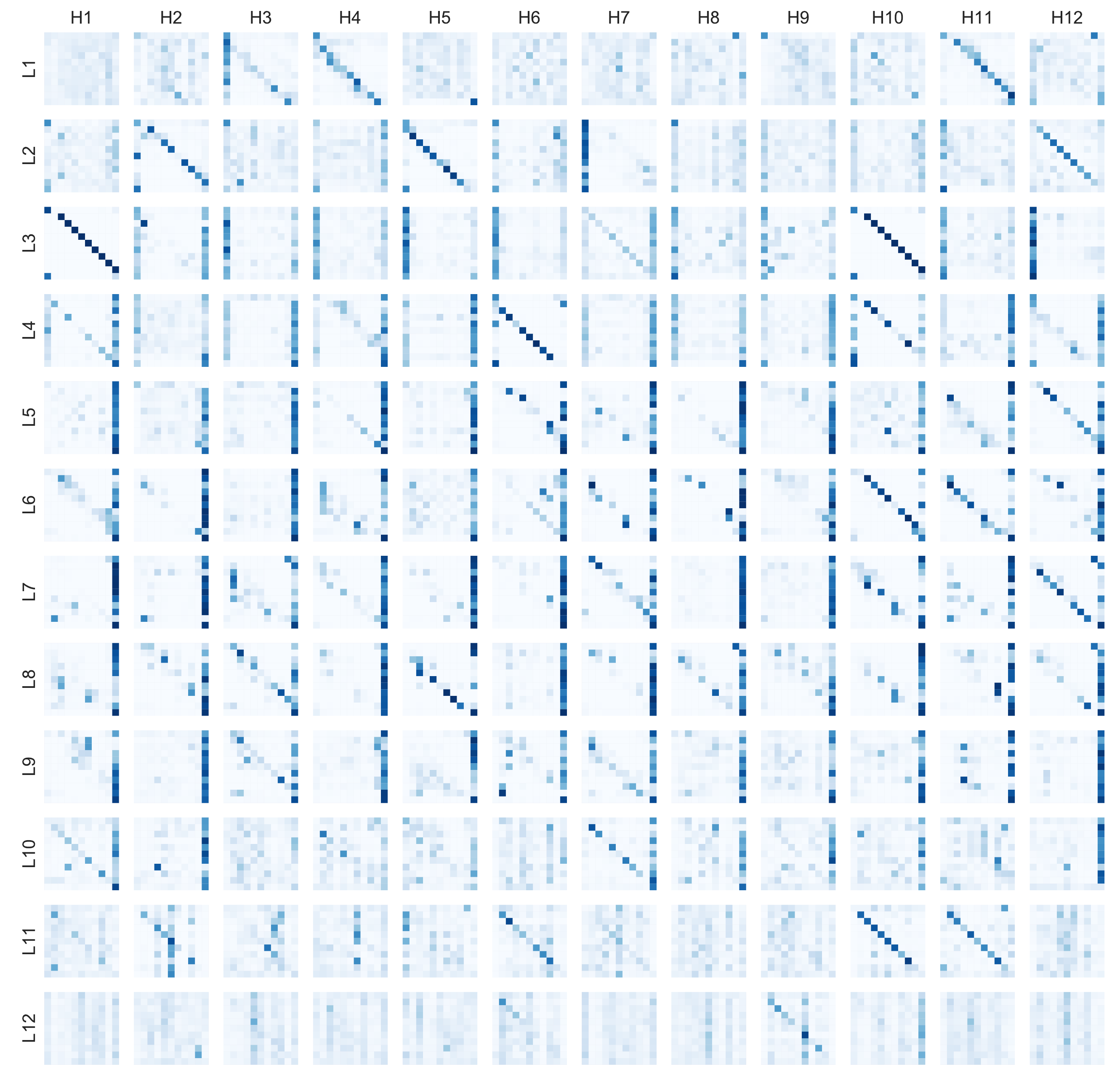}}  
	\subfloat[TernaryBERT.]{
		\includegraphics[width=0.5\textwidth]{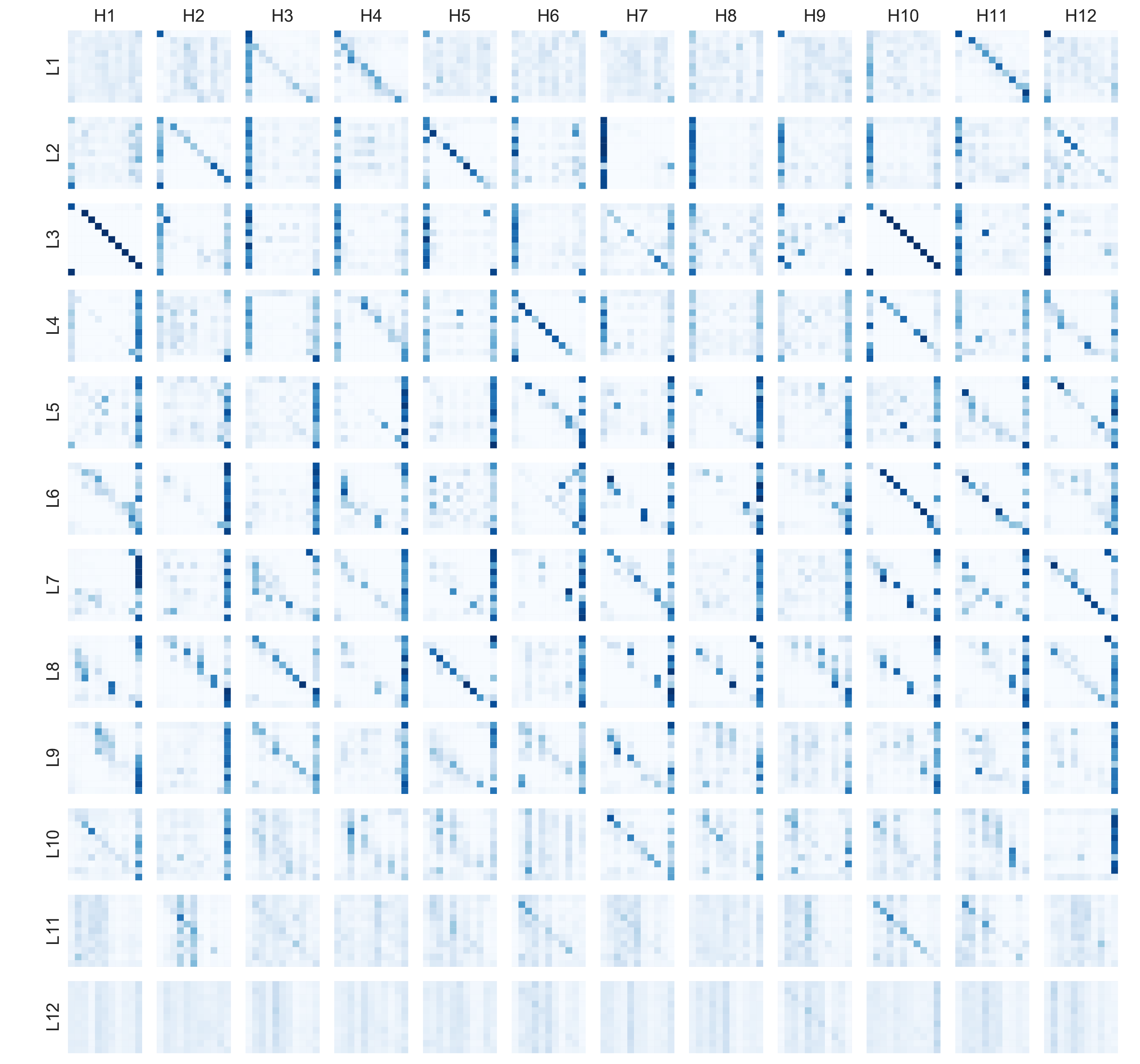}}
	\caption{Attention patterns of full-precision and ternary BERT trained on CoLA. The input sentence is ``Who does John visit Sally because he likes?''}
	\label{fig:cola1}
\end{figure*}

\begin{figure*}[htbp]
	\centering
	\subfloat[Full-precision BERT.]{
		\includegraphics[width=0.5\textwidth]{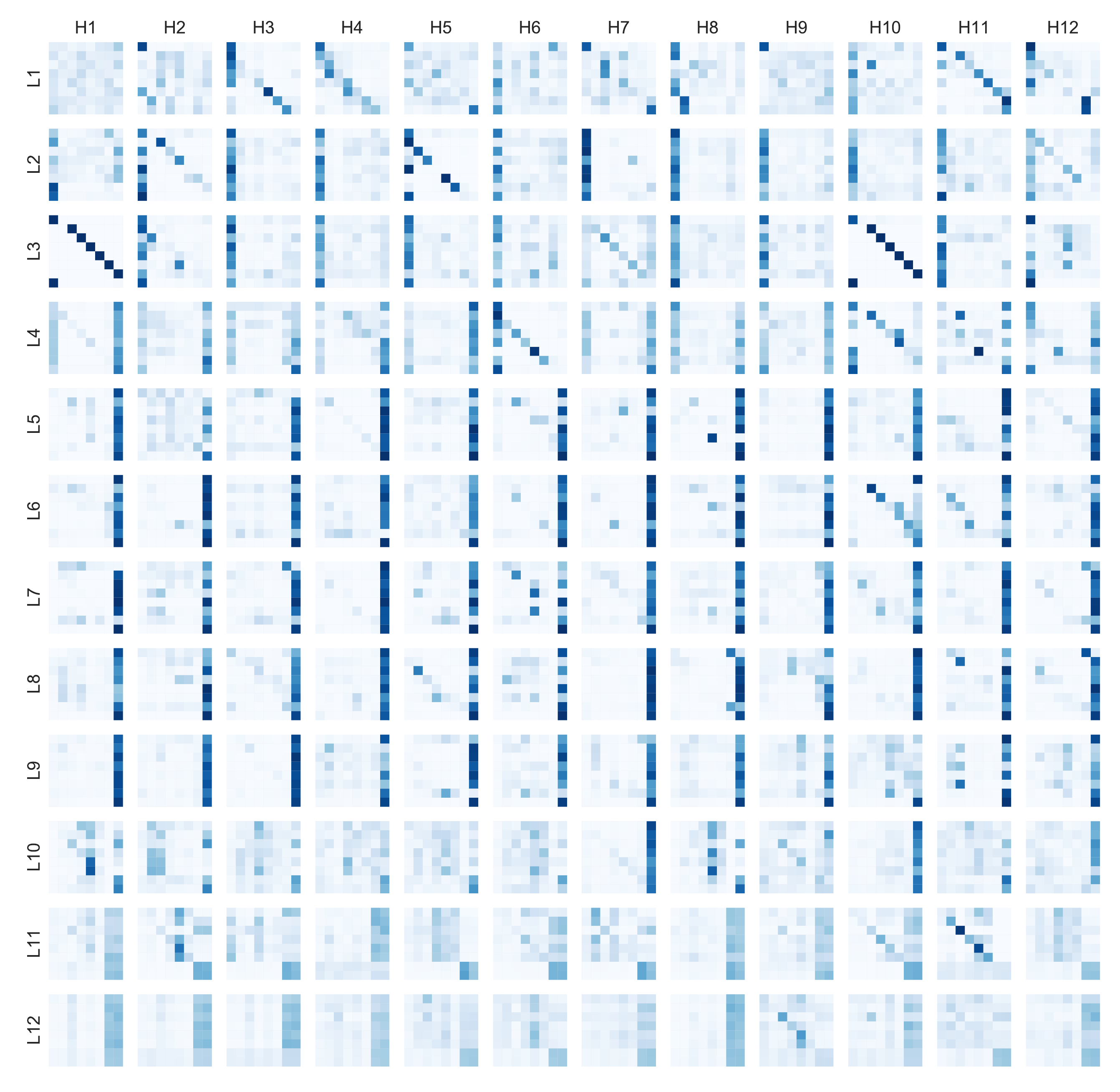}}
	\subfloat[TernaryBERT.]{
		\includegraphics[width=0.5\textwidth]{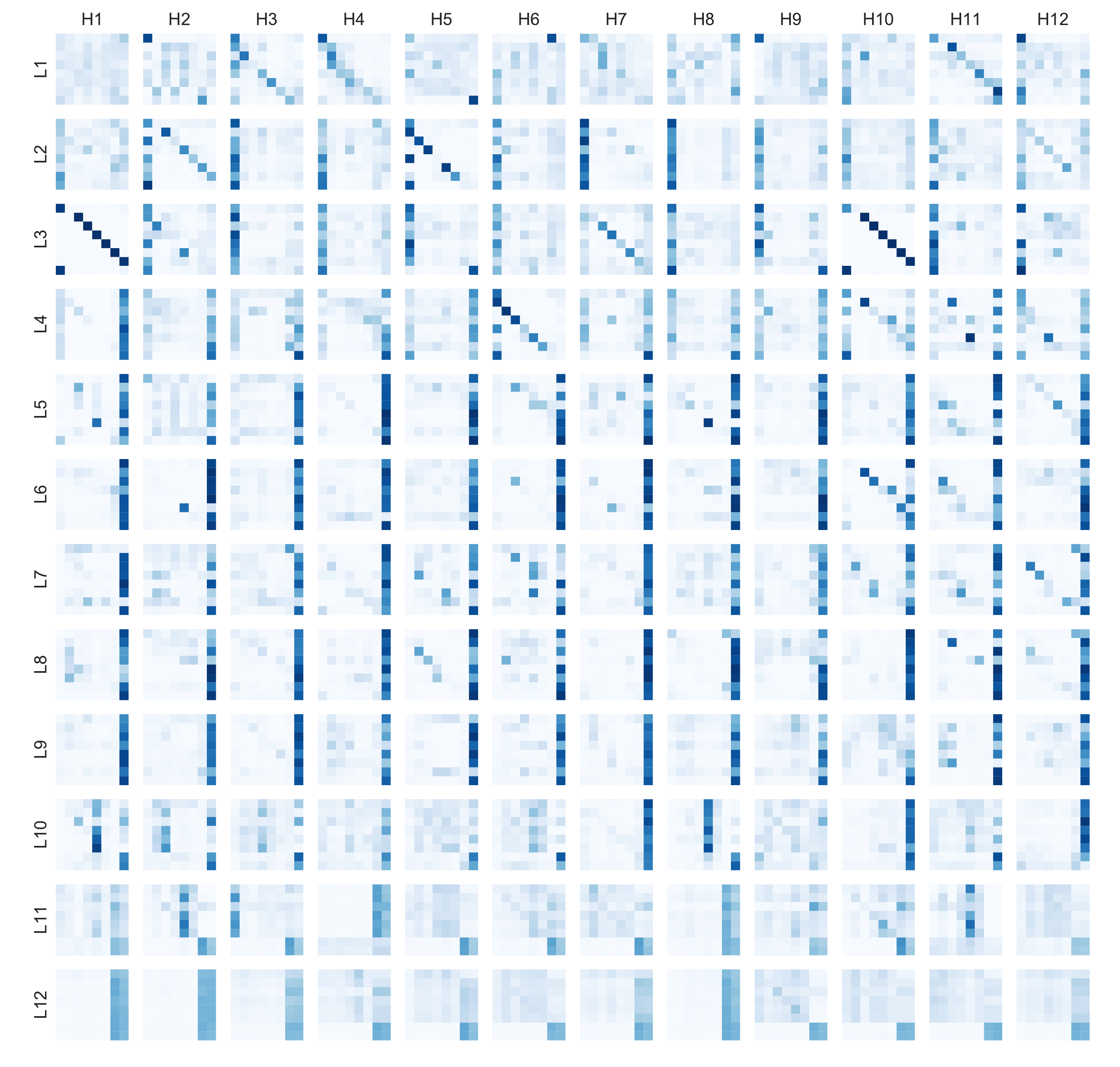}}
	\caption{Attention patterns of full-precision and ternary BERT trained on SST-2. The input sentence is ``this movie is maddening."}
	\label{fig:sst-2}
	\subfloat[Full-precision BERT.]{
		\includegraphics[width=0.5\textwidth]{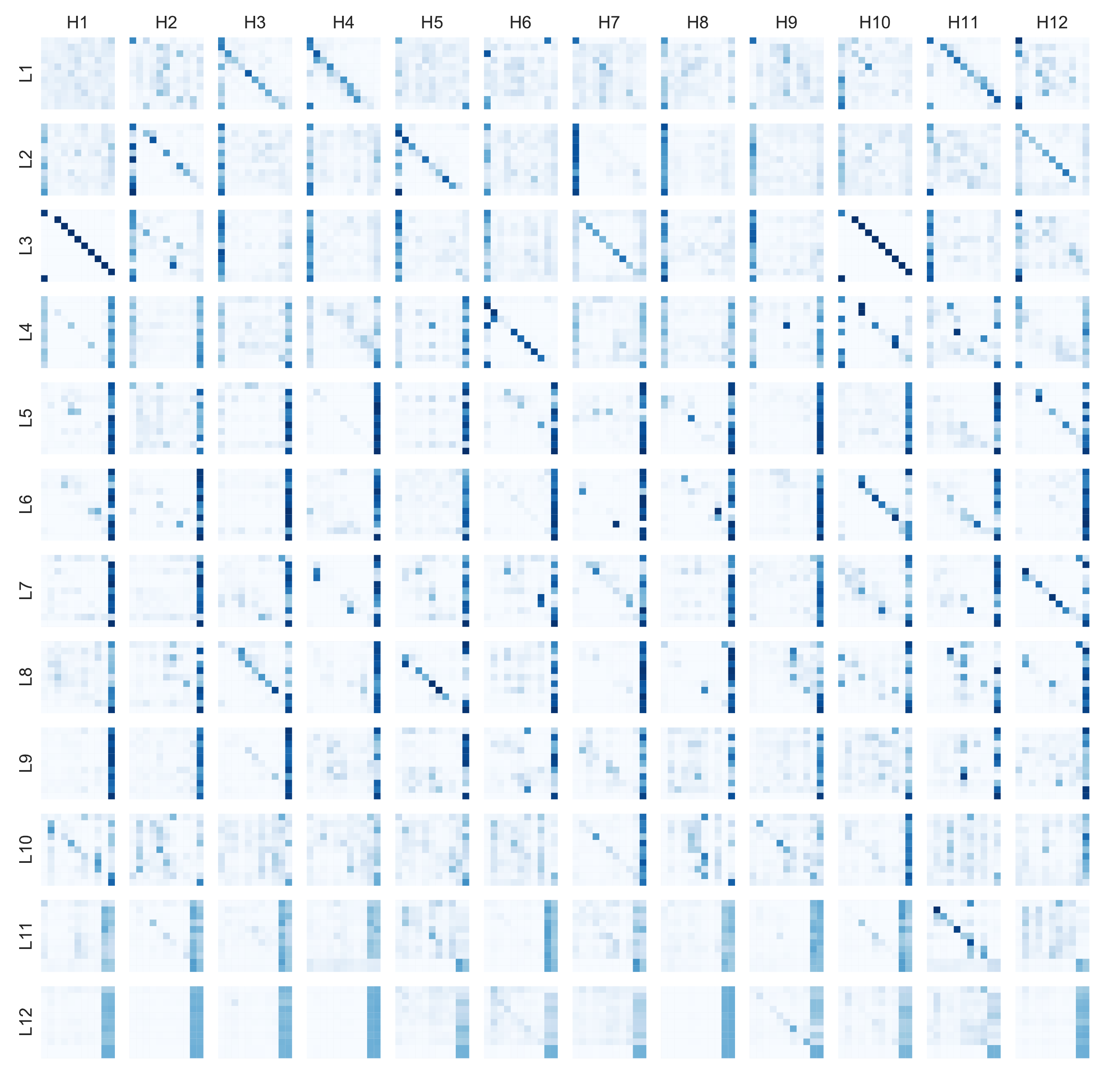}}
	\subfloat[TernaryBERT.]{
		\includegraphics[width=0.5\textwidth]{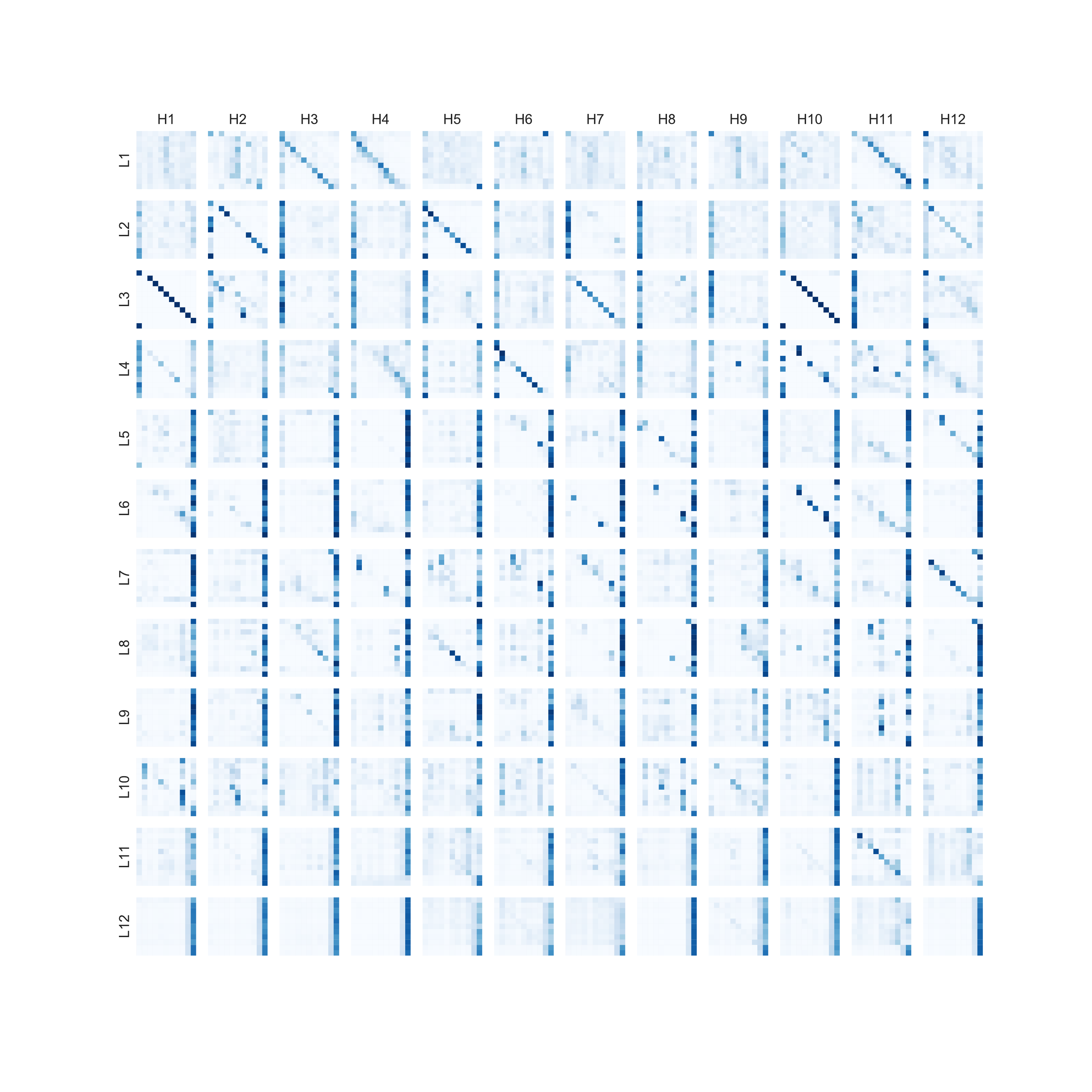}}
	\caption{Attention patterns of full-precision and ternary BERT trained on SST-2. The input sentence is ``old-form moviemaking at its best."}
	\label{fig:sst-21}
\end{figure*}

In Figures~\ref{fig:cola}-\ref{fig:sst-21}, we compare the attention  patterns of the fine-tuned full-precision BERT-base model and the ternarized $\text{TernaryBERT}_\text{TWN}$ on CoLA and SST-2. CoLA is a task which predicts the grammatical acceptability of a given sentence, and SST-2 is a task of classifying the polarity of movie reviews. 
As can be seen, the attention patterns of TernaryBERT resemble those in the full-precision BERT.

\end{document}